\documentclass[final]{cvpr}

\usepackage{times}
\usepackage{epsfig}
\usepackage{graphicx}
\usepackage{amsmath}
\usepackage{amssymb}
\usepackage{color}
\usepackage{epsfig}
\usepackage{graphicx}
\usepackage{cite}
\usepackage{multirow}
\usepackage{booktabs}
\usepackage{nth}
\usepackage{bbm}
\usepackage{wrapfig}
\usepackage[ruled,vlined]{algorithm2e}
\usepackage{caption}
\usepackage{subcaption}
\usepackage{bm}
\usepackage[title]{appendix}

\DeclareMathOperator*{\argmax}{argmax}

\definecolor{bingcolor}{RGB}{0, 160, 80}

\iftrue

\else
\fi

\usepackage[pagebackref=true,breaklinks=true,colorlinks,bookmarks=false]{hyperref}

\def\cvprPaperID{9904} 
\def\confYear{CVPR 2021}
\begin{document}

\title{SiamMOT: Siamese Multi-Object Tracking}

\author{ Bing Shuai \qquad
Andrew Berneshawi \qquad
Xinyu Li \qquad
Davide Modolo \qquad
Joseph Tighe \\

Amazon Web Services (AWS) \\
{\tt \small \{bshuai, bernea, xxnl, dmodolo, tighej\}@amazon.com}
}

\maketitle

\begin{abstract}

In this paper, we focus on improving online multi-object tracking (MOT). In particular, we introduce a region-based Siamese Multi-Object Tracking network, which we name SiamMOT. SiamMOT includes a motion model that estimates the instance's movement between two frames such that detected instances are associated. To explore how the motion modelling affects its tracking capability, we present two variants of Siamese tracker, one that implicitly models motion and one that models it explicitly. We carry out extensive quantitative experiments on three different MOT datasets: MOT17, TAO-person and Caltech Roadside Pedestrians, showing the importance of motion modelling for MOT and the ability of SiamMOT to substantially outperform the state-of-the-art. Finally, SiamMOT also outperforms the winners of ACM MM'20 HiEve Grand Challenge on HiEve dataset. Moreover, SiamMOT is efficient, and it runs at 17 FPS for 720P videos on a single modern GPU. 
Codes are available in \url{https://github.com/amazon-research/siam-mot}. 

\end{abstract}

\section{Introduction}
Multi-object tracking (MOT) is the problem of detecting object instances and then temporally associating them to form trajectories.  Early works~\cite{zhang2008global,berclaz2011multiple,zamir2012gmcp,kim2015multiple,tang2017multiple,henschel2017improvements,ristani2018features,sheng2018heterogeneous,xu2019spatial,xu2019train, wang2019exploit, andriyenko2011multi, berclaz2006robust, evangelidis2008parametric, tang2017multiple} formulate instance association as a graph-based optimization problem under the ``tracking-by-detection" paradigm, in which a node represents a detection and an edge encodes the likelihood of two nodes being linked. In practice, they use a combination of visual and motion cues to represent each node, which often requires expensive computation. Furthermore, they usually construct a large offline graph, which is non-trivial to solve, making them inapplicable for real-time tracking. 
Recently, online trackers \cite{bewley2016simple, wojke2017simple,bergmann2019tracking, zhou2020tracking} started to emerge, as they are more desirable in real-time tracking scenarios.  They focus on improving local linking over consecutive frames rather than building an offline graph to re-identify instances across large temporal gaps. Among these, some recent works \cite{bergmann2019tracking, zhou2019deep} have pushed \textit{online} MOT into state-of-the-art territory, making them very competitive.

In this work, we explore the importance of modelling motion in \textit{online MOT} by building upon ``Simple and Online Realtime Tracking'' (SORT) \cite{bewley2016simple, wojke2017simple} that underlies recent state-of-the-art models \cite{bergmann2019tracking, zhou2020tracking}.
%
%
In SORT, a better motion model is the key to improving its local linking accuracy. For example, SORT~\cite{bewley2016simple} uses Kalman Filters~\cite{kalman1960new} to model the instance's motion with simple geometric features, while the more recent state-of-the-art trackers~\cite{bergmann2019tracking,zhou2020tracking} learn a deep network to predict the displacement (motion) of instances based on both visual and geometric features, significantly outperforming the simpler SORT. 

We conduct our motion modelling exploration by leveraging a region-based Siamese Multi-Object Tracking network, which we name {\bf SiamMOT}. We combine a region-based detection network (Faster-RCNN \cite{ren2015faster}) with two motion models inspired by the literature on Siamese-based single-object tracking~\cite{li2018high, li2019siamrpn++, guo2020siamcar, bertinetto2016fully, fan2019siamese}: an implicit motion model (IMM) and an explicit motion model (EMM).
Differently from CenterTrack \cite{zhou2020tracking} that implicitly infers the motion of instances with point-based features \cite{duan2019centernet, tian2019fcos, qiu2020borderdet}, SiamMOT uses region-based features and develops (explicit) template matching to estimate instance motion, which is more robust to challenging tracking scenarios, such as fast motion.

We present extensive ablation analysis on three different multi-person tracking datasets. Our results suggest that instance-level motion modelling is of great importance for robust online MOT, especially in more challenging tracking scenarios. Furthermore, we show that the motion models of SiamMOT can improve tracking performance substantially, especially when cameras are moving fast and when people's poses are deforming significantly. 

On the popular MOT17 Challenge~\cite{milan2016mot16} SiamMOT with EMM achieves \textbf{65.9} MOTA  / \textbf{63.3} IDF1 with a DLA-34 \cite{yu2018deep} backbone by using \textit{public} detection, outperforming all previous methods. Moreover, on the recently introduced large-scale TAO-person dataset \cite{dave2020tao}, SiamMOT substantially improves over the state-of-the-art Tracktor++ \cite{bergmann2019tracking} from \textbf{36.7} to \textbf{41.1} TrackAP~\cite{dave2020tao, yang2019video}. Finally, we benchmark SiamMOT on the Human In Events (HiEve) dataset~\cite{lin2020human}, where it outperforms the winner of the ACM MM'20 grand HiEve challenge~\cite{lin2020hieve}. 

\section{Related work}

\subsection{Siamese trackers in SOT}

Single object tracking (SOT) refers to tracking a given object of interest, which is usually specified in the first frame and could belong to any semantic object class.    
Instead of detecting pre-defined objects in a frame and linking them back to earlier tracked instances, single object trackers (SOT) usually model the motion of the object of interest directly to predict its trajectory. 
Siamese-based trackers~\cite{held2016learning, bertinetto2016fully, li2018high, li2019siamrpn++, tao2016siamese, valmadre2017end, guo2017learning, he2018twofold, zhang2019structured, zhu2018distractor, fan2019siamese, zhang2019deeper, guo2020siamcar} are a family of state-of-the-art SOT. As the name suggests, Siamese trackers operate on pairs of frames. Their goal is to track (by matching) the target object in the first frame within a search region in the second frame. This matching function is usually learned offline on large-scale video and image datasets. 

In this paper, we formulate Siamese trackers within an end-to-end trainable multi-object tracking network (SiamMOT). The closest work to ours is DeepMOT that also trains Siamese trackers with other components under the proposed MOT training framework.  However, DeepMOT focuses on improving the structured loss in MOT rather than formulating the detector and tracker in a unified network, so an off-the-shelf single object tracker is needed in DeepMOT.  
Finally, while we take inspiration from particular Siamese trackers \cite{leal2016learning, li2018high, guo2020siamcar}, our formulation is generic enough that other Siamese trackers can easily be adapted in our MOT framework. 

\vspace{-1em}
\paragraph{Siamese network.} It's worth noting that Siamese trackers are different from general Siamese networks~\cite{leal2016learning, sun2019deep, varior2016siamese}. Siamese networks usually learn a affinity function between two detected instances, whereas Siamese trackers learn a matching function that is used to search for a detected instance within a larger contextual region.

\subsection{Tracking-by-Detection in MOT}
Many works tackle multi-object tracking (MOT) by adopting the ``tracking-by-detection" paradigm ~\cite{zhang2008global,berclaz2011multiple,zamir2012gmcp,kim2015multiple,tang2017multiple,henschel2017improvements,ristani2018features,sheng2018heterogeneous,xu2019spatial,xu2019train, sadeghian2017tracking, leal2016learning, wang2019exploit, andriyenko2011multi, berclaz2006robust, choi2010multiple, evangelidis2008parametric, fang2018recurrent}, where objects instances are first detected in each frame and then associated across time based on their visual coherence and spatial-temporal consistency. 
Some of these works focused on learning new functions to evaluate short-term associations more robustly \cite{ristani2018features, sheng2018heterogeneous, tang2017multiple, xu2019spatial, zhang2008global, tang2017multiple, sadeghian2017tracking, leal2016learning, choi2015near, fang2018recurrent, zhang2008global}. Others, instead, focused on learning how to output more temporally consistent long-term tracks by optimizing locally connected graphs~\cite{zhang2008global,berclaz2011multiple,zamir2012gmcp,kim2015multiple,tang2017multiple,henschel2017improvements,ristani2018features,sheng2018heterogeneous,xu2019spatial,xu2019train, wang2019exploit, andriyenko2011multi, berclaz2006robust, evangelidis2008parametric, tang2017multiple}. Many of these approaches are  inefficient, as they employ separate computationally expensive cues, like object detection~\cite{girshick2015fast, ren2015faster, dai2016r, he2017mask}, optical flow~\cite{dosovitskiy2015flownet, sun2018pwc, tang2017multiple, choi2015near}, and re-identification~\cite{hermans2017defense, tang2017multiple, zhou2019deep, tang2017multiple, ristani2018features}. 

\vspace{-1em}
\paragraph{Online MOT.} Online MOT refers to performing instance association on the fly without knowledge of future frames \cite{keuper2018motion, ban2016tracking, xiang2015learning, bewley2016simple, wojke2017simple, bergmann2019tracking, zhou2020tracking}. Therefore, online MOT focuses more on accurate local association rather than global-optimal association in which detections can be linked across long temporal gaps (as in offline graph modelling). It has seen a resurgence of popularity recently as new models are efficient enough to be applicable to real-time tracking.
For example, Ban et al.~\cite{ban2016tracking} formulated it in a probabilistic framework by using a variational expectation maximization algorithm to find the tracks. Xiang et al.~\cite{xiang2015learning} used Markov Decision Processes and reinforcement learning for online instance association. Bewley et al. \cite{bewley2016simple, wojke2017simple} developed simple object and realtime tracking (SORT) for quick online instance association.  SORT has been widely used in recent deep neural network based models \cite{zhou2020tracking, bergmann2019tracking} which achieve state-of-the-art performance on public MOT datasets. Our SiamMOT is based on SORT, and we explore how to improve its tracking performance. 

\vspace{-1em}
\paragraph{Motion modelling in SORT.} The original SORT~\cite{bewley2016simple} only used geometric features of tracks (location, box shape, etc) in its motion model to track  locations across frames. Later, Wojke et al.~\cite{wojke2017simple} improved SORT by incorporating visual features into the motion model to link the detected instances. Recently, Bergmann et al.~\cite{bergmann2019tracking} and Zhou et al.~\cite{zhou2020tracking} jointly learned the motion model with the detector such that both visual and geometric features are used.
In detail, Tracktor~\cite{bergmann2019tracking} leveraged a two stage detector~\cite{ren2015faster} to regress from previous person's location to current frame; CenterTrack~\cite{zhou2020tracking} adopted a track branch to regress the displacement of object centers between frames.
In this paper, we explore how to improve the motion model in SORT-based tracking model -- SiamMOT, and more importantly how it leads to improved MOT accuracy.


\begin{figure*}[t]
 \centering
  \includegraphics[width=1.0\textwidth]{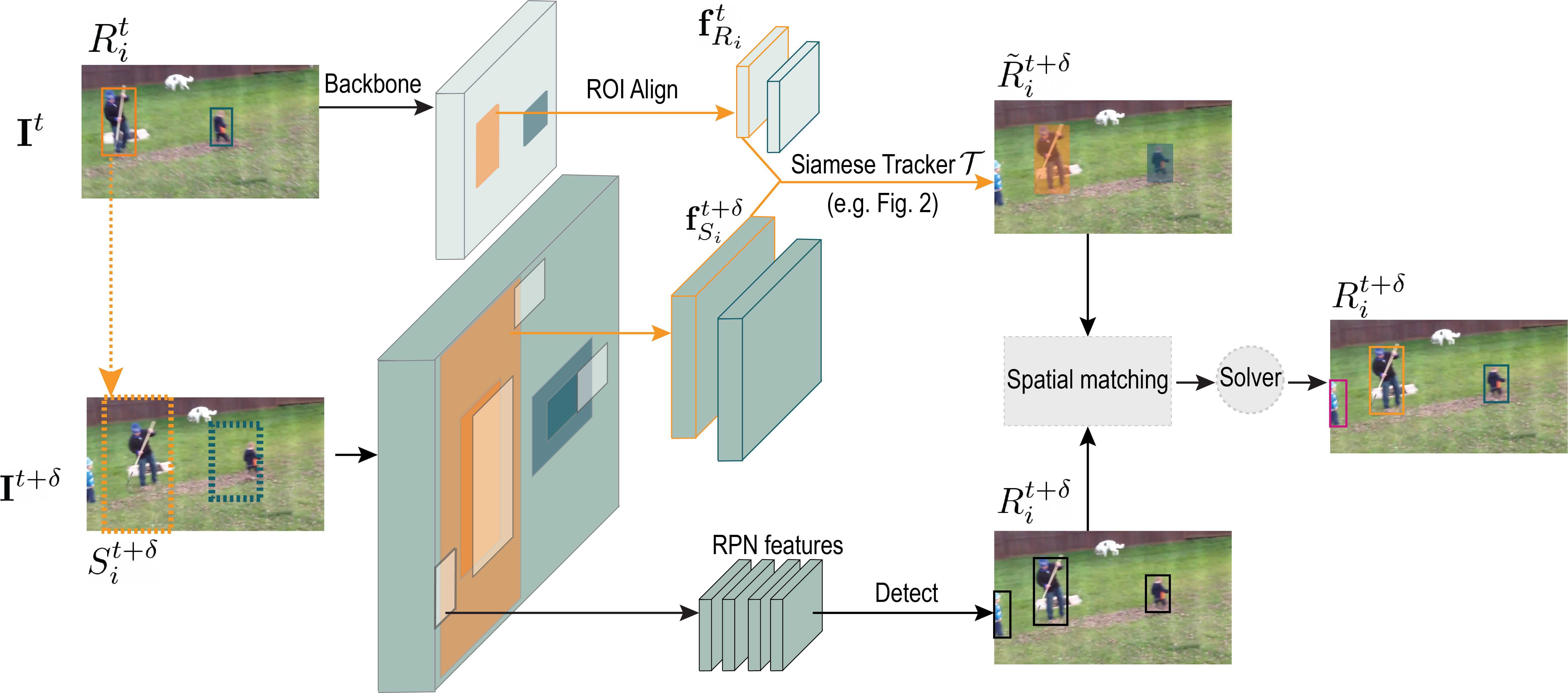}
  \caption{\small \it (Best viewed in color) SiamMOT is a region-based multi-object tracking network that detects and associates object instances simultaneously. The Siamese tracker models the motion of instances across frames and it is used to temporally link detection in online multi-object tracking. Backbone feature map for frame $\mathbf{I}^t$ is visualized with 1/2 of its actual size.  
  }
  \label{figure:teaser}
\end{figure*}

\section{SiamMOT: Siamese Multi-Object Tracking} \label{sec:architecture}
SiamMOT builds upon Faster-RCNN object detector~\cite{girshick2015fast, ren2015faster, he2017mask}, which consists of a Region Proposal Network (RPN) and a region-based detection network. On top of the standard Faster-RCNN, SiamMOT adds a region-based Siamese tracker to model instance-level motion. As shown in Fig.~\ref{figure:teaser}, SiamMOT takes as input two frames $\mathbf{I}^t, \mathbf{I}^{t+\delta}$ together with a set of detected instances $\mathbf{R}^t = \{R_1^t, \ldots R_i^t, \ldots \}$ at time $t$. In SiamMOT, the detection network outputs a set of detected instances $\mathbf{R}^{t+\delta}$, while the tracker propagates $\mathbf{R^t}$ to time $t+\delta$ to generate $\tilde{\mathbf{R}}^{t+\delta}$. 

As in SORT, SiamMOT contains a motion model that \textit{tracks} each detected instance from time $t$ to $t+\delta$ by propagating the bounding box $R_i^t$ at time $t$ to $\tilde{R}_i^{t+\delta}$ at $t+\delta$ ; and a spatial matching process that \textit{associates} the output of tracker $\tilde{R}_i^{t+\delta}$ with the detections $R_i^{t+\delta}$ at time ${t+\delta}$ such that detected instances are linked from $t$ to $t+\delta$. 

In the next section we introduce how our Siamese tracker models instance motion in SiamMOT (Sec.~\ref{sec:motion_model}) and present two variants of Siamese trackers in Sec.~\ref{sec:imm} and Sec.~\ref{sec:emm}. Finally, we provide the details for training and inference (Sec.~\ref{sec:train_infer}).

\subsection{Motion modelling with Siamese tracker}
\label{sec:motion_model}
In SiamMOT, given a detected instance $i$ at time $t$, the Siamese tracker searches for that particular instance at frame $\mathbf{I}^{t+\delta}$ in a contextual window around its location at frame $\mathbf{I}^t$ (i.e, $R_i^{t}$). Formally, 
\begin{equation}
     (v_i^{t+\delta}, \tilde{R}_i^{t+\delta}) = \mathcal{T}(\mathbf{f}_{R_i}^t, \mathbf{f}_{S_i}^{t+\delta}; \Theta)
\label{equation:sot}
\end{equation}
where $\mathcal{T}$ is the learnable Siamese tracker with parameters $\Theta$, $\mathbf{f}_{R_i}^t$ is the feature map extracted over region $R_i^{t}$ in frame $\mathbf{I}^{t}$, and $\mathbf{f}_{S_i}^{t+\delta}$ is the feature map extracted over the search region $S_i^{t+\delta}$ in frame $\mathbf{I}^{t+\delta}$. We compute $S_i^{t+\delta}$ by expanding $R_i^t$ by a factor r ($> 1$) while maintaining the same geometric center (e.g., dashed bounding box in Fig.~\ref{figure:teaser}). We extract features $\mathbf{f}_{R_i}^t$ and  $\mathbf{f}_{S_i}^{t+\delta}$ using the region of interest align (ROIAlign) layer of Mask-RCNN\cite{he2017mask}.
Finally, $v_i^{t+\delta}$ is the confidence of visibility for detected instance $i$ at time $t+\delta$. As long as the instance is visible in $S_i^{t+\delta}$, $\mathcal{T}$ should produce a high score $v_i^{t+\delta}$, otherwise $\mathcal{T}$ should produce a low score.  
Note how this formulation is reminiscent of that of Siamese-based single-object trackers~\cite{bertinetto2016fully, held2016learning, li2018high, li2019siamrpn++} and specifically, how they model the instance’s motion between frames.

In the context of multi-object tracking, we apply Eq.~\ref{equation:sot} multiple times, once for each detected instance $R_i^{t} \in \mathbf{R}^t$. Importantly, our SiamMOT architecture allows these operations to run in parallel and only requires the backbone features to be computed once, making online tracking inference efficient.

We conjecture that motion modelling is particularly important for online MOT.  
Specifically, association between $R^t$ and $R^{t+\delta}$ will fail if 1) $\tilde{R}^{t+\delta}$ does not match to the right instance in $R^{t+\delta}$ or 2) $v_i^{t+\delta}$ is low for a visible person at $t+\delta$. Previous works \cite{bergmann2019tracking, zhou2020tracking} approach the problem of regressing $\tilde{R}^{t+\delta}$ from the previous location (i.e. $R_i^t$) by feeding the model with features from both frames. By doing so these works aim to implicitly model the instance's motion in the network. 
However, as research in single-object tracking~\cite{li2018high, li2019siamrpn++, bertinetto2016fully, guo2020siamcar} reveals, finer-grained spatial-level supervision is of great significance to explicitly learn a robust target matching function in challenging scenarios.  Based on this rationale, we present two different parameterizations of $\mathcal{T}$ in SiamMOT -- an implict motion model in Sec.~\ref{sec:imm} and an explicit motion model in Sec.~\ref{sec:emm}.

\subsection{Implicit motion model}~\label{sec:imm}
Implicit motion model (IMM) uses an MLP to implicitly estimate the instance-level motion between two frames. 
In detail, the model concatenates $\mathbf{f}_{S_i}^t$ and $\mathbf{f}_{S_i}^{t+\delta}$ and feeds that to an MLP that predicts the visibility confidence $v_i$ and the relative location and scale changes:
\begin{equation}
m_i=[\frac{x_i^{t+\delta} - x_i^t }{w_i^t}, \ \frac{y_i^{t+\delta}-y_i^t}{h_i^t},  \text{log}\frac{w_i^{t+\delta}}{w_i^t} \ \text{log}\frac{h_i^{t+\delta}}{h_i^t}]
\label{equation:motion}
\end{equation}
in which $(x_i^t, y_i^t, w_i^t, h_i^t)$ is the parameterization of $R_i^t$. We can trivially derive $\tilde{R}^{t+\delta}$ from an inversion transformation of Eq.~\ref{equation:motion} by taking as input $R_i^t$ and $m_i$. 

\vspace{-1em}
\paragraph{Loss.} Given a triplet $(R_i^t, S_i^{t+\delta}, R_i^{t+\delta})$, we train IMM with the following training loss:
\begin{equation}
    \mathbf{L} = \ell_{focal}(v_i, v_i^*) + \mathbbm{1}[v_i^*] \, \ell_{reg}(m_i, m_i^*)
    \label{equation:imm}
\end{equation}
where $v_i^*$ and $m_i^*$ refer to ground truth values derived from $R_i^{t+\delta}$, $\mathbbm{1}$ is the indicator function, $\ell_{focal}$ the focal loss for classification~\cite{lin2017focal} and $\ell_{reg}$ the commonly used smooth $\ell_1$ loss for regression. Please refer to the supplementary material for the network architecture.

\subsection{Explicit motion model}~\label{sec:emm}

\begin{figure}[t]
     \centering
         \centering
         \includegraphics[width=0.5\textwidth]{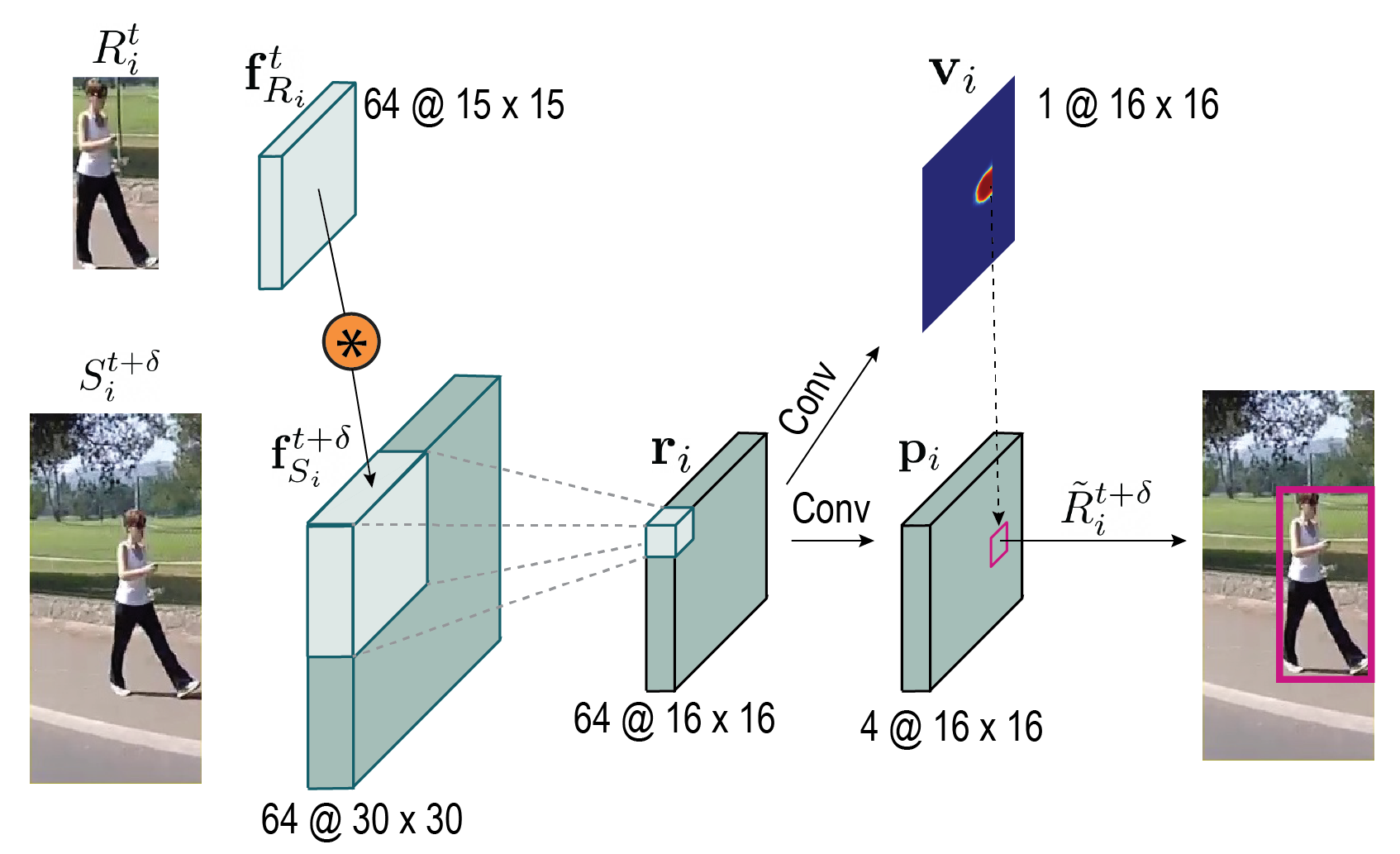}
        \caption{\small \it Network architecture of Explicit Motion Model (EMM), $*$ represents channel-wise cross correlation operator.}
        \label{fig:emm}
\end{figure}

Inspired by the literature on single-object tracking~\cite{li2018high, li2019siamrpn++, guo2020siamcar, tao2016siamese, bertinetto2016fully}, we propose an explicit motion model (EMM, Fig.\ref{fig:emm}) in SiamMOT. Specifically, it uses a channel-wise cross-correlation operator ($*$) to generate a pixel-level response map $\mathbf{r}_i$, which has shown to be effective in modelling dense optical flow estimation~\cite{dosovitskiy2015flownet} and in SOT for instance-level motion estimation~\cite{li2018high, li2019siamrpn++, bertinetto2016fully, guo2020siamcar}. In SiamMOT, this operation correlates each location of the search feature map $\mathbf{f}_{S_i}^{t+\delta}$ with the target feature map $\mathbf{f}_{R_i}^t$ to produce $\mathbf{r}_i = \mathbf{f}_{S_i}^{t+\delta} * \mathbf{f}_{R_i}^t$, so each map $\mathbf{r}_i[k, :, :]$ captures a different aspect of similarity. 
Inspired by FCOS~\cite{tian2019fcos}, EMM uses a fully convolutional network $\psi$ to detect the matched instances in $\mathbf{r}_i$. Specifically, $\psi$ predicts a dense visibility confidence map $\mathbf{v}_i$ indicating the likelihood of each pixel to contain the target object, and a dense location map $\mathbf{p}_i$ that encodes the offset from that location to the top-left and bottom-right bounding box corners. Thus, we can derive the instance region at $(x, y)$ by the following transformation $\mathcal{R}(\mathbf{p}(x,y)) = [x-l, y-t, x+r, y+b]$ in which $\mathbf{p}(x,y) = [l, t, r, b]$ (the top-left and bottom-right corner offsets).
Finally, we decode the maps as follows:
\begin{equation}
\begin{split}
    \tilde{R}_i^{t+\delta} = \mathcal{R}(\mathbf{p}_i(x^*, y^*)); \ \ \ v_i^{t+\delta} = \mathbf{v}_i(x^*, y^*)& \\
     \mathbf{s.t.} (x^*, y^*) = \argmax_{x,y}(\mathbf{v}_i \odot \bm{\eta}_i)& \\
\end{split}
\end{equation}
where $\odot$ is the element-wise multiplication, $\bm{\eta}_i$ is a penalty map that specifies a non-negative penalty score for the corresponding candidate region as follows:
\begin{equation}
    \bm{\eta}_i(x, y) = \lambda \mathcal{C} +  (1-\lambda)\mathcal{S}(\mathcal{R}(\mathbf{p}(x,y)), R_i^t)
\end{equation}
where $\lambda$ is a weighting scalar ($0\leq\lambda\leq1$), $\mathcal{C}$ is the cosine-window function w.r.t the geometric center of the previous target region $\mathcal{R}_i^t$ and $\mathcal{S}$ is a Guassian function w.r.t the relative scale (height / width) changes between the candidate region ($\mathbf{p}(x,y))$) and $ R_i^t$. The penalty map $\bm{\eta}_i$ is introduced to discourage dramatic movements during the course of tracking, similar to that in \cite{li2018high, li2019siamrpn++, guo2020siamcar, fan2019siamese}.

\vspace{-1em}
\paragraph{Loss.} Given a triplet $(R_i^t, S_i^{t+\delta}, R_i^{t+\delta})$, we formulate the training loss of EMM as follows:
\begin{equation}
\begin{split}
    \mathbf{L} &= \sum_{x, y}\ell_{focal}(\mathbf{v}_i(x,y), \mathbf{v}_i^*(x,y)) \\
               &+ \sum_{x, y}\mathbbm{1}[\mathbf{v}_i^*(x,y)=1](w(x,y) \cdot \ell_{reg}(\mathbf{p}_i(x,y), \mathbf{p}_i^*(x,y)))
\end{split}
\end{equation}
where $(x,y)$ enumerates all the valid locations in $S_i^{t+\delta}$, $\ell_{reg}$ is the IOU Loss for regression~\cite{yu2016unitbox, danelljan2019atom} and $\ell_{focal}$ is the focal loss for classification~\cite{lin2017focal}. 
Finally, $\mathbf{v}_i^*$ and $\mathbf{p}_i^*$ are the pixel-wise ground truth maps. $\mathbf{v}_i^*(x,y) = 1$ if $(x,y)$ is within $R_i^{*t+\delta}$ and $0$ otherwise. $\mathbf{p}_i^*(x,y) = [x-x^*_0, y-y^*_0, x^*_1-x, y^*_1-y]$ in which $(x^*_0, y^*_0)$ and $(x^*_1, y^*_1)$ corresponds to the coordinates of the top-left and the bottom-right corner of ground truth bounding box $R_i^{t+\delta}$. Similar to \cite{zhu2019soft}, we modulate $\ell_{reg}$ with $w(x,y)$, which is the centerness of location $(x, y)$ w.r.t to the target instance $R_i^{t+\delta}$ and is defined as $w(x, y) = \sqrt{\frac{\min(x-x_0, x_1-x)}{\max(x-x_0, x_1-x)} \cdot \frac{\min(y-y_0, y_1-y)}{\max(y-y_0, y_1-y)}}$.

EMM improves upon the IMM design in two ways. First it uses the channel independent correlation operation to allow the network to explicitly learn a matching function between the same instance in sequential frames. Second,
it enables a mechanism for finer-grained pixel-level supervision which is important to reduce the cases of falsely matching to distractors.

\subsection{Training and Inference}
\label{sec:train_infer}
We train SiamMOT in an end-to-end fashion with the following loss $\ell = \ell_{rpn} + \ell_{detect} + \ell_{motion}$, in which  $\ell_{rpn}$  and  $\ell_{detect}$ are the standard losses for RPN~\cite{ren2015faster} and the detection sub-network~\cite{girshick2015fast} in Faster-RCNN. $\ell_{motion} = \sum_{x_i \in \mathcal{X}} \mathbf{L}(x_i)$ is used to train the Siamese tracker, wherein $\mathcal{X} = \cup_{i=1}^M (R_i^t, S_i^{t+\delta}, R_i^{t+\delta})$ are training triplets.  Note that $R_i^{t+\delta} = \varnothing$ if $R_i^t$ does not include a ground truth instance or the instance in $R_i^t$ is not visible in $S_i^{t+\delta}$. Similar to Faster-RCNN training, we sample $R_i^t$ from the outputs of the RPN~\cite{ren2015faster}.

At inference, a standard IOU-based NMS operation is first used on the outputs of the detection sub-network ($R^{t+\delta}$ in Fig.~\ref{figure:teaser}) and on those of the Siamese tracker ($\tilde{R}^{t+\delta}$ in Fig.~\ref{figure:teaser}) independently. Next, the following spatial matching process is used to merge $R^{t+\delta}$ and $\tilde{R}^{t+\delta}$: detections that spatially match ($IOU \geq 0.5$) to any tracked instance are suppressed and thus removed.
Then, we adopt a standard \textit{online} solver as that in \cite{bergmann2019tracking,zhou2020tracking,bewley2016simple,wojke2017simple}: 1) a trajectory is continued if its visibility confidence ($v_i^t$) is above $\alpha$; 2) a trajectory is born if there is a non-matched detection and its confidence is above $\beta$ and 3) a trajectory is killed if its visibility confidence ($v_i^t$) is below $\alpha$ for consecutive $\tau$ frames.

\vspace{-1em}
\paragraph{Short occlusion handling.} In the case of short occlusions, the visibility confidence for the target would be low (lower than the threshold $\alpha$). Instead of killing them, we keep those tracks in memory and continue searching for them in future frames (up to $\tau > 1$ frames) to check whether they can be re-instated. We use the last predicted location and its corresponding feature as the searching template.

\section{Experimental settings}

\subsection{Datasets and Metrics} \label{sec:datasets}
\noindent {\bf MOT17}\cite{milan2016mot16} is the most widely used multi-person tracking benchmark. It consists of 7 training and 7 test videos, ranging from from $7$ to $90$ seconds long. The videos feature crowded scenes in indoor shopping malls or outdoor streets.
We follow the evaluation protocol of~\cite{milan2016mot16} and report our results using several metrics: MOTA (Multiple Object Tracking Accuracy), IDF1 (ID F1 score), FP (False Positives), FN (False Negatives) and IDsw (ID switches). \\

\begin{figure}
     \centering
     \begin{subfigure}[b]{0.245\textwidth}
         \centering
         \includegraphics[width=1.0\textwidth]{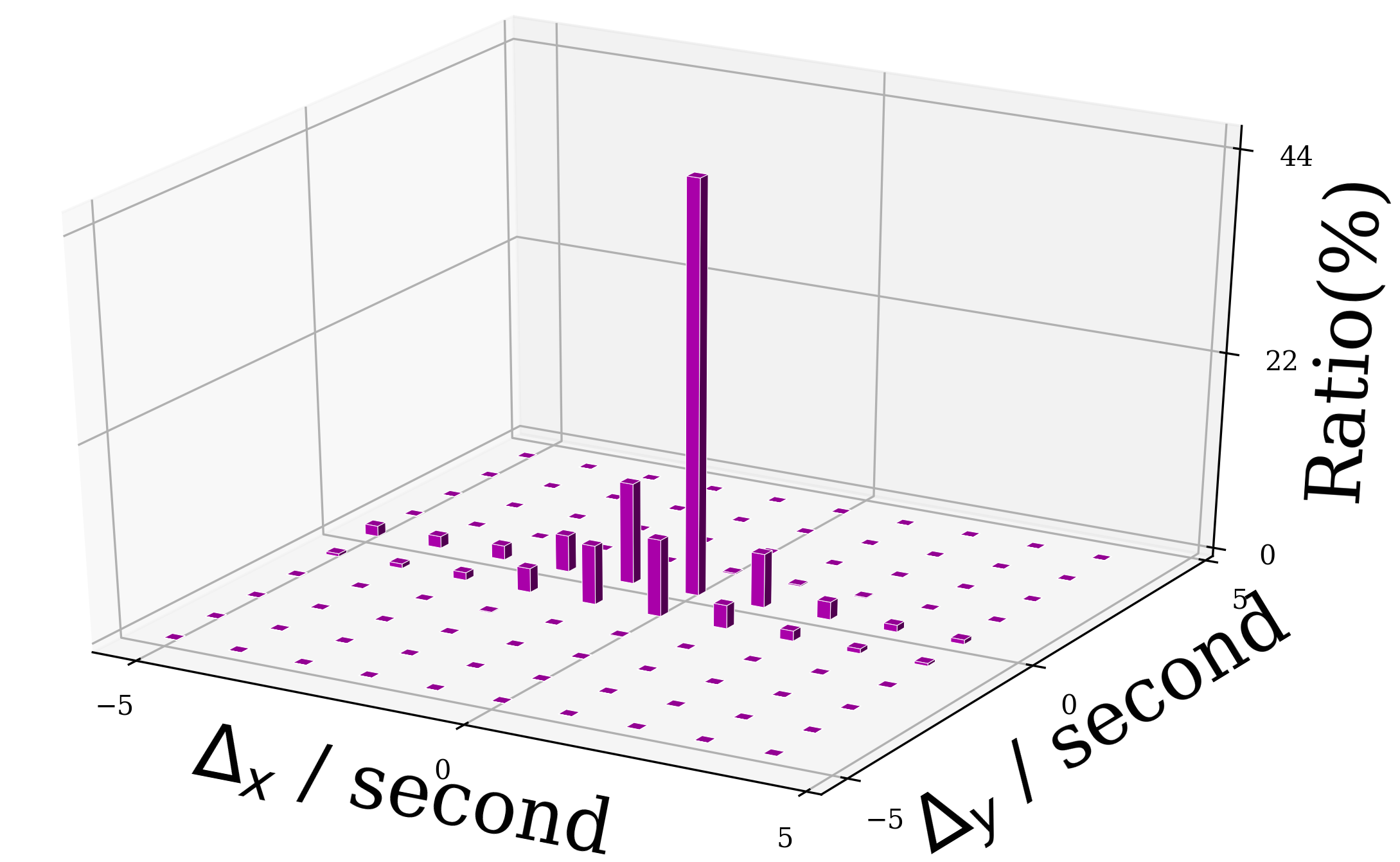}
         \caption{\small \it MOT17}
     \end{subfigure}
     \begin{subfigure}[b]{0.245\textwidth}
         \centering
         \includegraphics[width=1.0\textwidth]{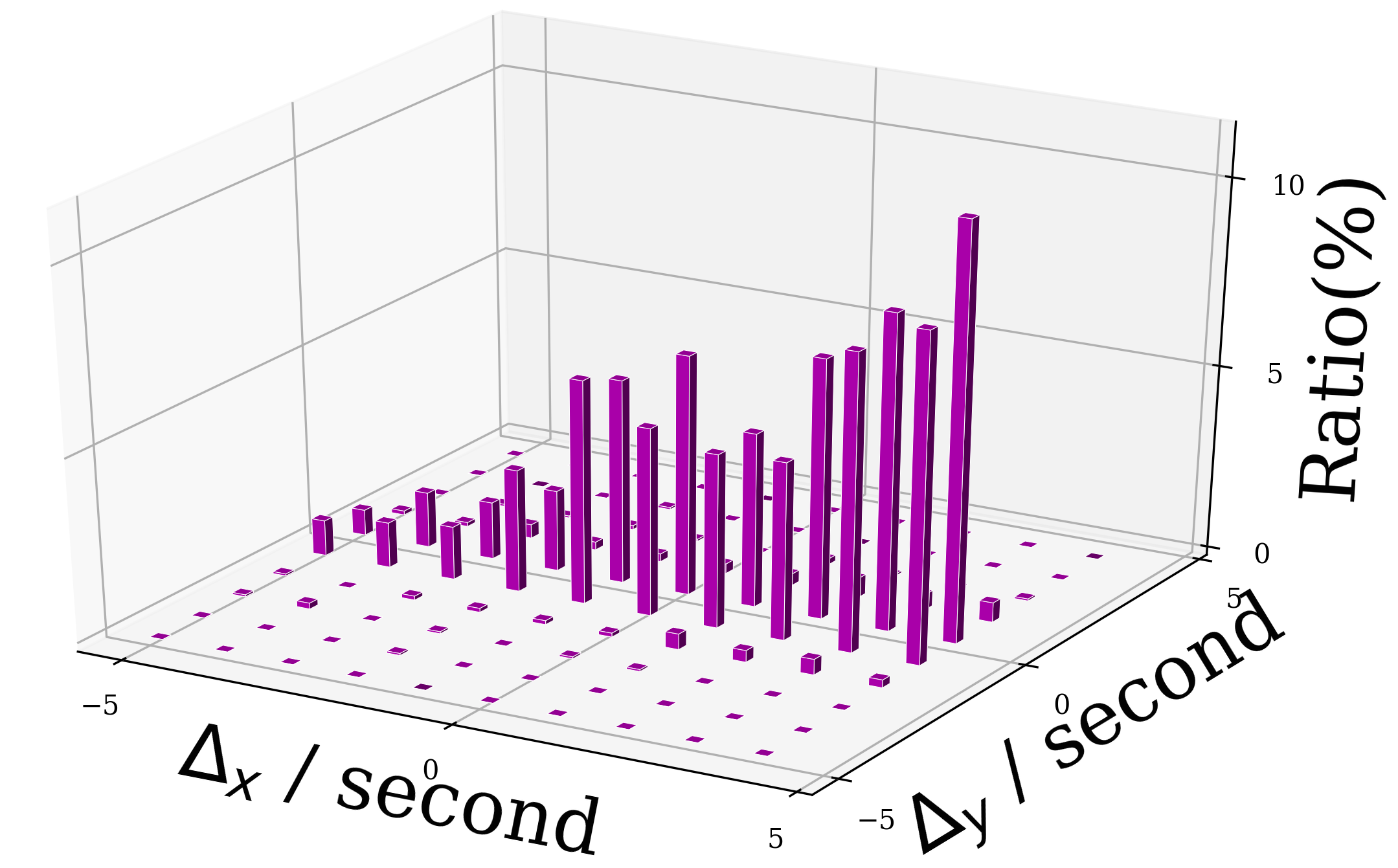}
         \caption{\small \it CRP}
     \end{subfigure}
     \begin{subfigure}[b]{0.245\textwidth}
         \centering
         \includegraphics[width=1.0\textwidth]{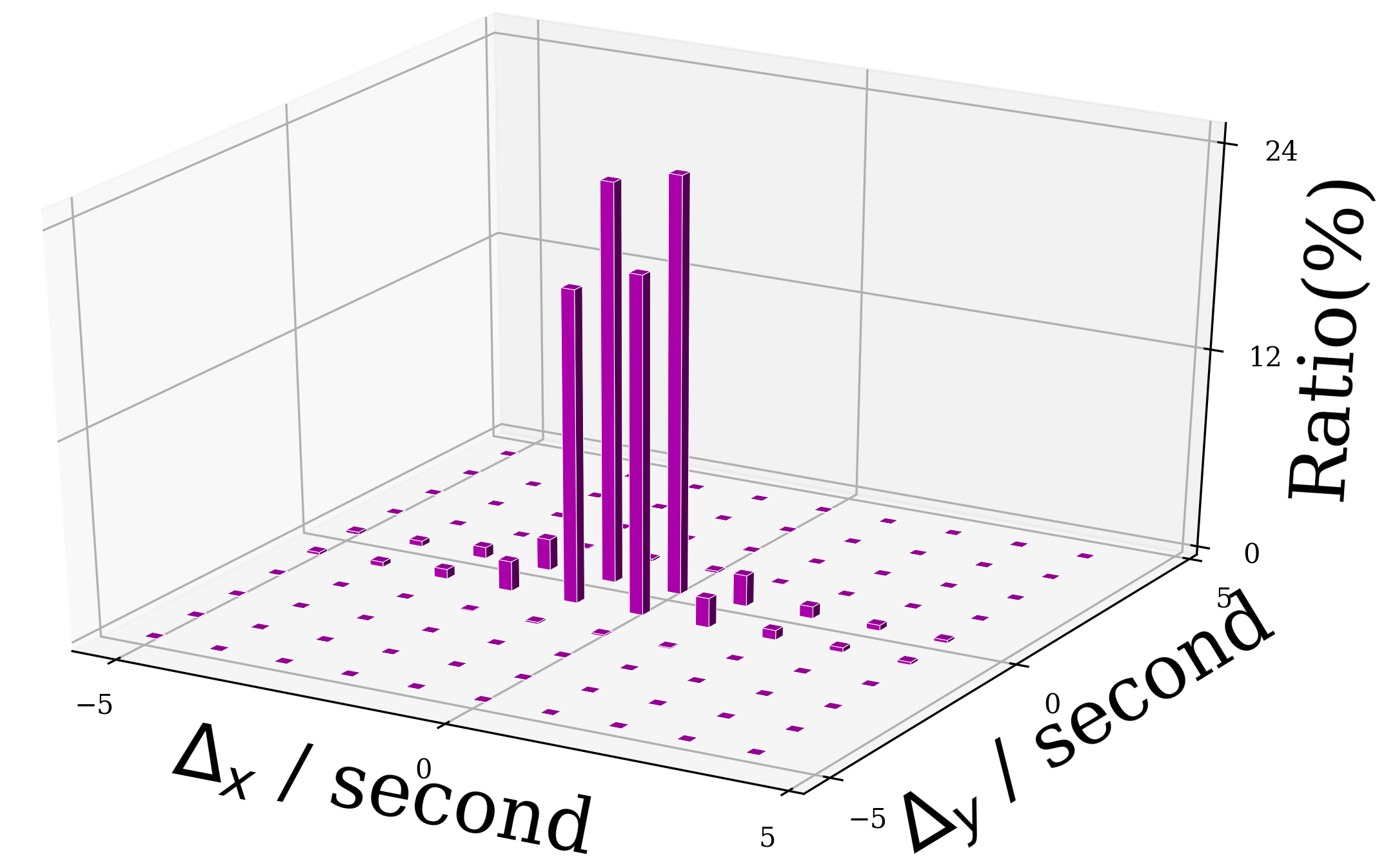}
         \caption{\small \it TAO-Person}
     \end{subfigure}
        \caption{\small \it 3D histogram of normalized motion offset per second across different datasets. 
        }
        \label{fig:dataset_motion_stat}
\end{figure}

\vspace{-1em}
\noindent {\bf TAO-person}\cite{dave2020tao} is a newly-established large scale multi-person tracking benchmark. It is a subset of the TAO dataset~\cite{dave2020tao} and it consists of 418 training and 826 validation videos. To include a large variability of scenes, the videos are collected by mixing existing datasets like AVA~\cite{gu2018ava} (generic movies), Charades~\cite{sigurdsson2016hollywood} (indoor activities), BDD~\cite{yu2020bdd100k} (streets), Argoverse~\cite{Argoverse} (streets) and other sports videos. This dataset contains rich motion artifacts (e.g. motion and defocus blur), as well as diverse person motion patterns (Fig.~\ref{fig:dataset_motion_stat}{\color{red}c}), which makes tracking persons challenging. 
We follow the evaluation protocol of~\cite{dave2020tao} and use the provided toolbox to report Federated Track-AP (TAP). Federated evaluation \cite{gupta2019lvis} is used because not all videos are exhaustively annotated.  Different from MOTA, Track-AP \cite{yang2019video} highlights the temporal consistency of the underlying trajectories.\\

\vspace{-1em}
\noindent {\bf Caltech Roadside Pedestrians (CRP)}\cite{hall2015fine} is a dataset for person analysis in videos. It consists of 7 videos, each roughly 20 minutes long. The videos are captured from a camera mounted to a car while driving, and they mainly feature outdoor roadside scenes. Due to the fast camera motion, the pedestrians appear as they are moving relatively much faster than in other datasets (Fig.~\ref{fig:dataset_motion_stat}{\color{red}b}). We report results on the same metrics used for MOT17.\\

\vspace{-1em}
\noindent {\bf Datasets analysis.} 
Each of these datasets contains different challenges for tracking. For example, tracking people in MOT17 is challenging due to occlusion and crowded scenes, even though people do not move fast and their poses are constant (i.e., standing). In contrast, scenes in CRP are not as crowded, but the camera motion is very large and the pedestrian's position changes quickly. Finally, TAO includes a wide range of scene types and video corruption artifacts.
As we focus on modelling short term motion for tracking, here we examine the characteristics of motion in each of these datasets. Towards this, we calculate the ground truth motion vector $\mathbf{m}$ as in Eq.~\ref{equation:imm} for every person, between two consecutive annotated frames. As videos are not annotated densely (i.e., every frame), we normalize  $\mathbf{m}$ by $\delta$ (their time difference). We present dataset-specific histograms in Fig.~\ref{fig:dataset_motion_stat}. People in MOT17 dataset have relatively small motion compared to those in TAO and CRP.

\subsection{Implementation details}

\paragraph{Network.} We use a standard DLA-34~\cite{yu2018deep} with feature pyramid~\cite{lin2017feature} as the Faster-RCNN backbone. We set $r = 2$, so that our search region is $2 \times$ the size of the tracking target. In IMM, $\mathbf{f}_{S_i}^t$ and  $\mathbf{f}_{S_i}^{t+\delta}$ have the same shape $\mathbb{R}^{c \times 15 \times 15}$ and the model is parametrized as a 2-layer MLP with $512$ hidden neurons. In EMM, instead, $\mathbf{f}_{R_i}^{t} \in \mathbb{R}^{c \times 15 \times 15}$ and $\mathbf{f}_{S_i}^{t+\delta} \in \mathbb{R}^{c \times 30 \times 30}$, so that they are at the same spatial scale; the model is a 2-layer fully convolutional network, with stacks of $3 \times 3$ convolution kernels and group normalization~\cite{wu2018group}). 

\vspace{-1em}
\paragraph{Training samples.} As previously mentioned, we train SiamMOT on pairs of images. When video annotations are not available, we follow ~\cite{held2016learning, zhou2020tracking} by employing \textit{image training}, in which spatial transformation (crop and re-scale) and video-mimicked transformation (motion blur) are applied to an image such that a corresponding image pair is generated. When video annotations are available, we use \textit{video training}, in which we sample pairs of two random frames that are at most 1 second apart.

\vspace{-1em}
\paragraph{Training.} We jointly train the tracker and detection network. We sample 256 image regions from the output of the RPN to train them. We use SGD with momentum as the optimizer, and we train our model for $25K$ and $50K$ iterations, for CrowdHumand \cite{shao2018crowdhuman} and COCO \cite{lin2014microsoft} datasets respectively. We resize the image pair during training such that its shorter side has 800 pixels. We start training with a learning rate of $0.02$ and decrease it by factor $10$ after $60\%$ of the iterations, and again after $80\%$. We use a fixed weight decay of $10^{-4}$ and a batch size of $16$ image pairs.

\vspace{-1em}
\paragraph{Inference.} We empirically set linking confidence $\alpha = 0.4$ and detection confidence $\beta= 0.6$, and we present the sensitivity analysis of $\alpha$ and $\beta$ in the supplementary material. We keep a trajectory active until it is unseen for $\tau=30$ frames.

\begin{table*}[t]
\resizebox{\linewidth}{!}{
\centering
    \begin{tabular}{l|c|ccccc|ccccc | c}
    \toprule
    \multicolumn{1}{l}{Models} & \multicolumn{1}{c}{} & \multicolumn{5}{c}{MOT17} & \multicolumn{5}{c}{Caltech Roadside Pedestrians (CRP)} & \multicolumn{1}{c}{TAO-person} \\
    \toprule
    & Runtime   & MOTA $\uparrow$ & IDF1 $\uparrow$ & FP $\downarrow$ & FN $\downarrow$ & IDsw $\downarrow$ & MOTA $\uparrow$ & IDF1 $\uparrow$ & FP $\downarrow$ & FN $\downarrow$ & IDsw $\downarrow$ & TAP@0.5 $\uparrow$ \\
    \midrule
    Faster-RCNN (Tracktor) & 23.0 fps & 58.6 & 53.0 & 3195 & 42488 & 858 & 15.9 & 25.1 & 632 & 21238 & 1126 & 29.1\%\\
    Faster-RCNN + Flow  & 12.5 fps  & 60.3 & 54.8 & 3518 & 40387 & 716 & 41.8 & 56.4 & 2381 & 11934 & 1594 & 32.8\% \\
    Faster-RCNN + IMM & 19.5 fps & 61.5 & 57.5 & 5730 & 36863 & 678 & 76.8 & 81.2 & 2583 & 2391 & 1377 & 34.7\% \\
    Faster-RCNN + EMM & 17.6 fps & 63.3 & 58.4 & 5726 & 34833 & 671 & 76.4 & 81.1 & 2548 & 2575 & 1311 & 35.3\% \\
    \bottomrule
    \end{tabular}
}
\caption{\small \it Results on MOT17 train, Caltech Roadside Pedestrians and TAO-Person datasets. FPS are calculated based on MOT17 videos that are resized to 720P. IMM and EMM are the motion model presented for SiamMOT.}
\label{table:ablation_mot_crp_tao}
\end{table*}

\section{Ablation analysis} ~\label{sec:ablation}
We carry out ablation analysis on MOT17, CRP and TAO-person, which are considerably different from each other (sec.~\ref{sec:datasets}, Fig.~\ref{fig:dataset_motion_stat}) and provide a good set for ablation study. 
We adopt \textit{image training} to train SiamMOT, as we don't have large-scale video annotations to train a generalized model. Specifically, we train models from the full-body annotation of CrowdHuman \cite{shao2018crowdhuman}, and evaluate it on MOT17-train and CRP datasets as they have amodal bounding box annotation. We train models from visible-body annotations from CrowdHuman and COCO \cite{lin2014microsoft} and evaluate it on the TAO-person dataset. We do this to try to keep the models as comparable as possible while still adhering to the annotation paradigm of each dataset (amodal vs modal person bounding boxes). In order to directly compare frame-to-frame tracking, we adopt the same solver as that in Tracktor\cite{bergmann2019tracking}, in which the trajectory is killed immediately if it is unseen (i.e. $\tau=1$ frame).

\subsection{Instance-level motion modelling} \label{sec:ilmm}
We investigate the benefits of motion modelling for MOT (Table~\ref{table:ablation_mot_crp_tao}). 
We compare SiamMOT with IMM and EMM against two baselines: 
(1) our implementation of $\operatorname{Tracktor}$~\cite{bergmann2019tracking}, which we obtain by removing the Siamese tracker from SiamMOT and instead use the detection network to regress the location of the target in the current frame, and 
(2) $\operatorname{Tracktor+Flow}$, that adds a flow-based model to estimate the movement of people across frames.  This flow-based model can be considered a simple forward tracker that ``moves'' the previous target region to the current frame and then uses the detection network (as in $\operatorname{Tracktor}$) to regress to its exact location in the current frame. The movement of the person instance is estimated by taking the median flow field of its constituent pixels.
In our experiments we use a pre-trained state-of-the-art PWC-net~\cite{sun2018pwc} to estimate the pixel-wise optical flow field. 
Finally, for fair comparison we use the same detections for all four models. 

Results show that our implementation of $\operatorname{Tracktor}$ achieves competitive results on both MOT17 and TAO-person (higher than those reported by \cite{dave2020tao},\cite{bergmann2019tracking}), but performs poorly on CRP, as its motion model is too weak to track people that move too fast.
Adding flow to $\operatorname{Tracktor}$ significantly improves its performance ($\operatorname{Tracktor+Flow}$), especially on the challenging CRP and TAO-person datasets. 
$\operatorname{SiamMOT}$ improves these results even further, for both $\operatorname{IMM}$ and $\operatorname{EMM}$. The performance gap is especially interesting on the CRP dataset, where both MOTA and IDF1 increased substantially (i.e., $+35$ MOTA and $+25$ IDF1 over $\operatorname{Tracktor+Flow}$).
Between these, $\operatorname{EMM}$ performs similar to $\operatorname{IMM}$ on CRP, but significantly better on MOT17 and TAO-person. This shows the importance of explicit template matching, which is consistent with what observed in the SOT literature~\cite{li2019siamrpn++, leal2016learning}.
Finally, note that tracking performance keeps increasing as we employ better motion models (i.e., $\operatorname{Tracktor} < \operatorname{Flow} < \operatorname{IMM} < \operatorname{EMM}$). This further validates the importance of instance-level motion modelling in MOT. In addition, SiamMOT are significantly more efficient than $\operatorname{Tracktor+Flow}$, in which flow does not share computation with $\operatorname{Tracktor}$.

\subsection{Training of SiamMOT: triplets sampling}
\begin{table}[t]
\resizebox{\columnwidth}{!}{
    \centering
    \begin{tabular}{l|c @{\hskip 0.5em} c @{\hskip 0.5em} c @{\hskip 0.5em} c @{\hskip 0.5em} c @{\hskip 0.5em} |c}
    \toprule
    \multicolumn{1}{c}{Sampled triplets} & \multicolumn{5}{c}{MOT17} & \multicolumn{1}{c}{TAO-person}\\
    \toprule
     & MOTA $\uparrow$ & IDF1 $\uparrow$ & FP $\downarrow$ & FN $\downarrow$ & IDsw $\downarrow$ & TAP@0.5 $\uparrow$\\
     \midrule
     P + H & 59.7 & 58.6 & 9639 & 34976 & 618 & 34.2\% \\
     P + N & 62.7 & 58.3 & 6275 & 34955 & 697 & 35.0\% \\
     P + H + N & 63.3 & 58.4 & 5726 & 34833 & 671 & 35.3\% \\
     \bottomrule
    \end{tabular}
   }
    \caption{\small \it Effects of sampled triplets for training forward tracker in SiamMOT. P / N / H are positive / negative / hard training triplet. P+H triplets are usually used in single-object tracking.
    }
    \label{table:ablation_sampling}
\end{table}

We now evaluate how the distribution of triplets used to train SiamMOT (sec.~\ref{sec:train_infer}) affects its tracking performance.
Given a set of training triplets $\mathcal{X} = \cup_{i=1}^N (R_i^t, S_i^{t+\delta}, R_i^{t+\delta})$ from image pair $\{\mathbf{I}^t, \mathbf{I}^{t+\delta}\}$, a triplet can be either negative, positive or hard. It is negative (N) when $R_i^t$ does not include a person, positive (P) when $R_i^t$ and $S_i^{t+\delta}$ includes the same person, and hard (H) when $R_i^t$ includes a person, but $S_i^{t+\delta}$ does not include the target person.

Similar to the training of SOT, we start by training the Siamese tracker with positive and hard negative ($\operatorname{P+H}$) triplets.  As results in Tab.~\ref{table:ablation_sampling} shows, the model achieves reasonable IDF1 on MOT17, which means that the tracker can follow a true person quite robustly, but it achieves relatively low MOTA, as it occasionally fails to kill false positive tracks. This is because the Siamese tracker in SiamMOT usually starts with noisy detection rather than with human-annotated regions (as in SOT). Instead, $\operatorname{P+N}$ performs better and combining all of them $\operatorname{P+H+N}$ achieves the best results overall.

\subsection{Training of SiamMOT: joint training}
We now investigate the importance of training the region-based detection network jointly with our Siamese tracker. First, we look at the impact that joint training has on the accuracy of our tracker and later on the accuracy of the person detector.

\noindent {\bf Tracking performance.} 
We train a model only with the Siamese tracker (i.e. detection branch is discarded) and utilize the same detections used in the experiments presented in sec.~\ref{sec:ilmm} and Tab.~\ref{table:ablation_mot_crp_tao}. The MOTA achieved by EMM on MOT17 is 63.3 with joint training vs 61.5 without.  This gap shows the benefits of joint training.

\noindent {\bf Detection performance.} We compare two Faster-RCNN models trained with and without our Siamese tracker on MOT17. These models achieve 73.3\% and 73.4\%  AP@IOU=$0.5$, which indicates that the joint training in SiamMOT has no negative impact on the detection network.

Overall, these results show that joint training is very important for SiamMOT and leads to the best results. 

\subsection{Inference of SiamMOT}

\begin{table}[t]
\resizebox{\linewidth}{!}{
    \centering
    \begin{tabular}{l|c @{\hskip 0.5em} c @{\hskip 0.5em} c @{\hskip 0.5em} c @{\hskip 0.5em} c @{\hskip 0.5em} |c}
    \toprule
    \multicolumn{1}{c}{$\tau$ (frames)} & \multicolumn{5}{c}{MOT17} & \multicolumn{1}{c}{TAO-person}\\
    \toprule
     & MOTA $\uparrow$ & IDF1 $\uparrow$ & FP $\downarrow$ & FN $\downarrow$ & IDsw $\downarrow$ & TAP@0.5 $\uparrow$\\
     \midrule
     1  & 63.3 & 58.4 & 5726 & 34833 & 671 & 35.3\% \\
     5 & 63.5 & 60.0 & 5818 & 34497 & 622 & 35.6\% \\
     15 & 63.4 & 60.8 & 5979 & 34454 & 616 & 36.3\% \\
     30 & 63.3 & 60.6 & 6106 & 34465 & 658 & 36.6\% \\
     60 & 63.0 & 60.2 & 6510 & 34385 & 699 & 37.2\% \\
     \bottomrule
    \end{tabular}
    }
    \caption{\small \it Results of SiamMOT inference that terminates active trajectories after they are unseen within $\tau$ consecutive frames.}
    \label{table:ablation_track_inference}
\end{table}

Finally, we investigate how the inference of SiamMOT affects MOT performance. Similar to Tracktor \cite{bergmann2019tracking} and CenterTrack \cite{zhou2020tracking}, SiamMOT focuses on improving local tracking as long as the person is visible. However, a person can be shortly invisible due to occlusion (e.g. when crossing each other) that is common in crowded scenes such as MOT17. In order to track through these cases, we allow SiamMOT to track forward even when the trajectory is not visible, i.e. the tracker does not terminate the trajectory until it fails to track the corresponding target for $\tau$ consecutive frames. Results in Tab.~\ref{table:ablation_track_inference} show that tracking performance increases with $\tau$, especially IDF1 score / TrackAP that measures the temporal consistency of trajectories. This means that our tracker is capable of tracking beyond few consecutive frames. Results also show that the improvement saturates around $\tau=30$ (1s for 30 FPS videos). The reason is that people have likely moved outside of our search region by that time. In the future we will explore improving the motion modelling of the tracker in SiamMOT such that it can track through longer occlusions. 

\begin{table}[t]
\resizebox{\columnwidth}{!}{
	\centering
		\begin{tabular}{l| @{\hskip 0.5em}  c |@{\hskip 0.5em}  c @{\hskip 0.5em} c @{\hskip 0.5em} c @{\hskip 0.5em} c @{\hskip 0.5em} c @{\hskip 0.5em} c @{\hskip 0.5em} c }
			\toprule
			Method   & MOTA  & IDF1 & MT  & ML  & FP  & FN  & IDsw  \\
			\midrule
			STRN \cite{xu2019spatial}  & 50.9 & 56.5 & 20.1\% & 37.0\% & 27532 & 246924 & 2593 \\
			Tracktor++ \cite{bergmann2019tracking}   & 53.5 & 52.3 & 19.5\% & 36.6\% & 12201 & 248047 & 2072 \\
			DeepMOT \cite{xu2019train}   & 53.7 & 53.8 & 19.4\% & 36.6\% & 11731 & 247447 & 1947 \\
			Tracktor++ v2 \cite{bergmann2019tracking}  & 56.5 & 55.1 & 21.1\% & 35.3\% & 8866 & 235449 & 3763 \\
			NeuralSolver \cite{braso2020learning}  & 58.8 & 61.7 & 28.8\% & 33.5\% & 17413 & 213594 & 1185 \\
			CenterTrack\cite{zhou2020tracking}  & 61.5 & 59.6 & 26.4\% & 31.9\% & 14076 & 200672 & 2583\\
			\midrule
			SiamMOT  & \textbf{65.9} & \textbf{63.3} & 34.6\% & 23.9\% & 18098 & 170955 & 3040 \\
			\bottomrule
		\end{tabular}
	}
	\vspace{-2mm}
	\caption{\small \it  Results on MOT17 test set with public detection.} 
	\label{table:mot17_results}
\end{table}

\section{Comparison to State-of-the-art}
\label{sec:results}

Finally, we compare our SiamMOT with state-of-the-art models on three challenging multi-person tracking datasets: MOT17 \cite{milan2016mot16}, TAO-person \cite{dave2020tao} and HiEve Challenge \cite{lin2020human}.

\paragraph{MOT17} (Tab.~\ref{table:mot17_results}). We report results on the test set using publicly released detections, as done for the official MOT17 Challenge. We use EMM as the tracker of SiamMOT, pre-train using {\it image training} on CrowdHuman and train on MOT17 using {\it video training}.

We obtain our results by submitting SiamMOT predictions to the official evaluation server of the challenge\footnote{\url{https://motchallenge.net/}}. The results show that SiamMOT outperforms all previous methods, including the popular Tracktor++ v2 (+9.4 MOTA) and state-of-the-art CenterTrack \cite{zhou2020tracking} (+4.4 MOTA). 

Note how SiamMOT models instance's motion with region-based features while CenterTrack uses point-based features. As recent research shows \cite{qiu2020borderdet, yang2019reppoints, tian2019directpose}, region-based features are consistently better for instance recognition and localization. We conjecture this is also true for instance tracking. In addition, CenterTrack implicitly learns to infer the instance's motion in a similar way to the proposed IMM, which is not as good as EMM, as shown in Sec.~\ref{sec:ablation}, and by a large body of research in single-object tracking~\cite{leal2016learning, li2019siamrpn++, guo2020siamcar, tao2016siamese}.

 \begin{table}[t]
 \centering
\resizebox{0.8\linewidth}{!}{
	\centering
		\begin{tabular}{l|@{\hskip 0.5em} l@{\hskip 0.5em} c@{\hskip 0.5em} c }
			\toprule
			Method  & Backbone   & TAP@0.5  & TAP@0.75   \\
			\midrule
			Tractor \cite{dave2020tao} & ResNet-101  & 26.0\% & n/a \\
			Tracktor++ \cite{dave2020tao} & ResNet-101  & 36.7\% & n/a \\
			SiamMOT & ResNet-101 &  \textbf{41.1}\% & \textbf{23.0}\% \\
			\midrule
			SiamMOT & DLA-169 &  42.1\% & 24.3\% \\
			SiamMOT+ & DLA-169 & \textbf{44.3}\% & \textbf{26.2}\% \\
			\bottomrule
		\end{tabular}
	}
	\caption{\small \it Results on TAO-person validation set.}
	\label{table:tao_results}
\end{table}

\vspace{-1em}
\paragraph{TAO-person} (Tab.~\ref{table:tao_results}). We report results on the validation set similar to \cite{dave2020tao}. We train a SiamMOT with EMM using {\it image training} on MSCOCO and CrowdHuman datasets.  SiamMOT outperforms state-of-the-art Tracktor++ by a significant 4.4\ TrackAP@0.5.
As pointed out in \cite{dave2020tao}, linking tracklets with person re-identification embeddings is important in the TAO dataset, as there are a number of videos where people are moving in and out of camera view, which, in this case, is beyond the capability of instance-level motion modelling. Thus, we evaluate SiamMOT+ that merges tracklets with an off-the-shelf person re-id model, the one used in Tracktor++ \cite{bergmann2019tracking}. Thanks to this, SiamMOT+ sets new state-of-the-arts on the challenging TAO-person dataset. Although Tracktor++ gains a large 8\% TrackAP@0.5 boost by re-id linking, we observe a less significant improvement for SiamMOT. This is because our motion model is already capable of linking challenging cases in TAO, reducing the cases where re-id linking is necessary.

\begin{table}[t]
\resizebox{\linewidth}{!}{
	\centering
		\begin{tabular}{l @{\hskip 0.5em} | c @{\hskip 0.5em} c @{\hskip 0.5em} c @{\hskip 0.5em} c @{\hskip 0.5em} c @{\hskip 0.5em} c @{\hskip 0.5em} c }
			\toprule
			Method  & MOTA & IDF1  & MT  & ML  & FP  & FN  & IDsw  \\
			\midrule
			DeepSORT \cite{wojke2017simple} & 27.1 & 28.6 & 8.5\% & 41.5\% & 5894 & 42668 & 2220 \\
			FCS-Track \cite{lin2020hieve} & 47.8 & 49.8 & 25.3\% & 30.2\% &	3847 & 30862 & 1658 \\
			Selective JDE \cite{wu2020transductive} & 50.6 & 56.8 & 25.1\% & 30.3\% & 2860	& 29850	& 1719 \\
			LinkBox \cite{peng2020dense} & 51.4 & 47.2 & 29.3\% & 29.0\% & 2804 & 29345 & 1725 \\
			\midrule
			SiamMOT (DLA-34)  & 51.5 & 47.9 & 25.8\% & 26.1\% & 3509 & 28667 & 1866 \\
			SiamMOT (DLA-169)  & \textbf{53.2} & 51.7 & 26.7\% & 27.5\% & 2837 & \textbf{28485} & 1730 \\
			\bottomrule
		\end{tabular}
	}
	\caption{\small \it HiEve benchmark leaderboard (public detection). }
	\label{table:hie_results}
\end{table}

\paragraph{HiEve challenge} (Tab.~\ref{table:hie_results}).
Finally, to further show the strength of SiamMOT, we present results on the recently released Human in Events (HiEve) dataset~\cite{lin2020human}, hosted at the HiEve Challenge at ACM MM'20 ~\cite{lin2020hieve}. 
The dataset consists of 19 training and 13 test videos with duration ranging from $30$ to $150$ seconds, and the videos mainly feature surveillence scenes in subways, restaurants, shopping malls and outdoor streets.
We report results on the test set using the publicly released detections. We jointly train a SiamMOT with EMM on CrowdHuman and HiEve training videos. We obtain our results by submitting its predictions to the official evaluation server of the challenge\footnote{\url{http://humaninevents.org/}}. We submit two sets of results, one obtained with a lightweight DLA-34 backbone and one with a heavier-weight DLA-169. While the former already matches the top performance in ACM MM'20 HiEve Challenge \cite{lin2020hieve}, the latter beats all winning methods that are heavily tuned for the challenge. 

\section{Conclusion}

We presented a region-based MOT network -- SiamMOT, which detects and associates object instances simultaneously. 
In SiamMOT, detected instances are temporally linked by a Siamese tracker that models instance motion across frames. We found that the capability of the tracker within SiamMOT is particularly important to the MOT performance. We applied SiamMOT to three different multi-person tracking datasets, and it achieved top results on all of them, demonstrating that SiamMOT is a state-of-the-art tracking network. Although SiamMOT has proven to work well on person tracking, its framework can be easily adapted to accommodate multi-class multi-object tracking, and we plan to explore this direction in the future.

\begin{appendices}
\section{Implicit Motion Model}
\begin{figure}[htp]
    \centering
    \includegraphics[width=0.45\textwidth]{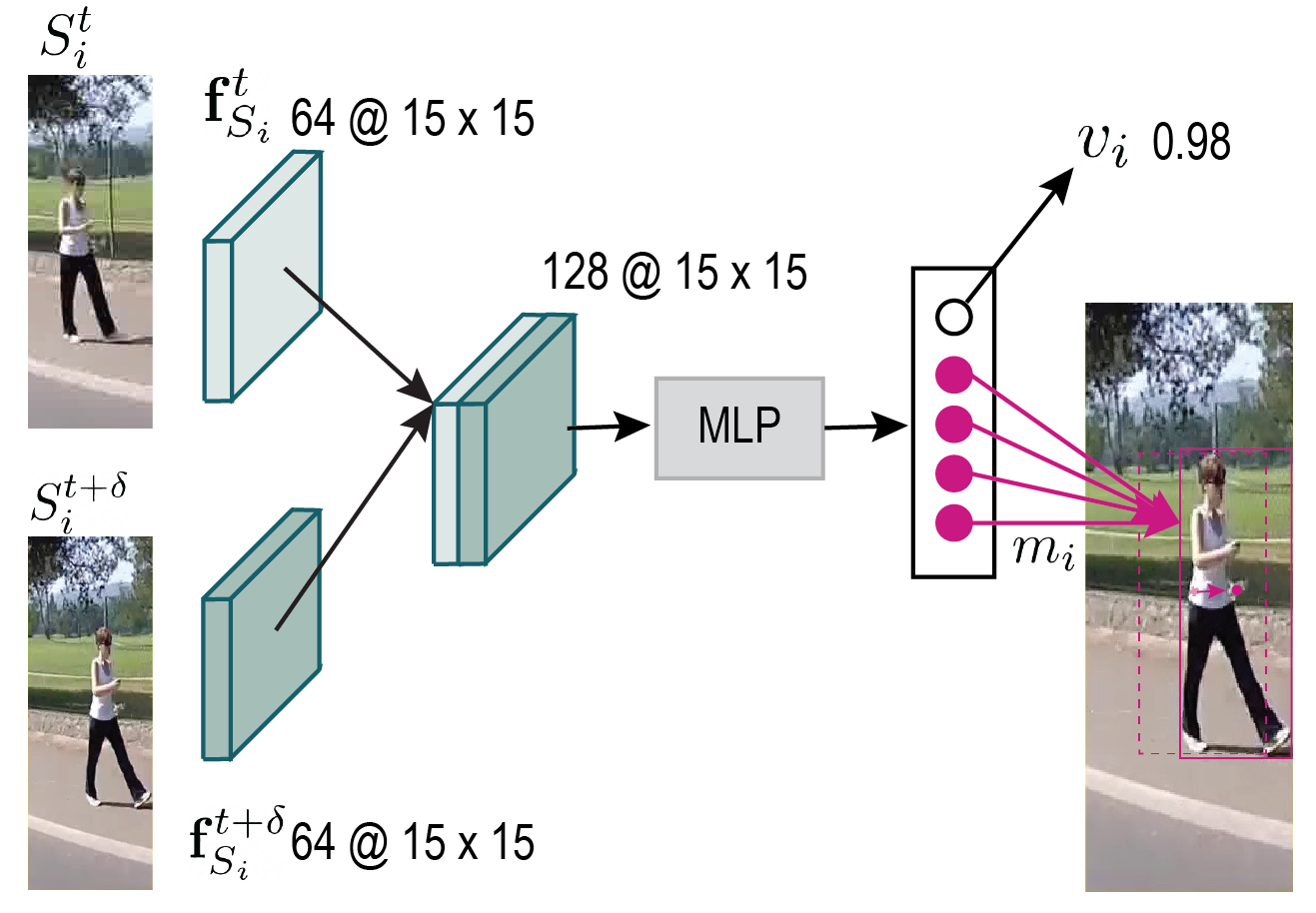}
    \caption{\small \it Network architecture of Implicit Motion Model (IMM). }
    \label{fig:imm}
\end{figure}

We show the graphic illustration of our Implicit Motion Model (IMM) in Fig. \ref{fig:imm}. Please refer to the main paper for definition of mathematical notation. In general, IMM learns the relative location / scale changes (encoded in $m_i$) of person instances with visual features of both frames. We empirically set the shape of $\mathbf{f}_{S_i}^{t+\delta}$ to be $c \times 15 \times 15 $, and we observe diminished performance gain when we increase it to $c \times 30 \times 30 $. Under current configurations, IMM has already entailed significantly more ($400 \times$) learnable parameters than EMM in the parameterization of Siamese tracker.

\section{Explicit Motion Model}
During inference, we empirically set $\lambda=0.4$ in generating penalty map ($\bm \eta_i$) by default. Due to the large person motion in CRP videos, we use $\lambda = 0.1$, which does not heavily penalize a matched candidate region if it is far away from the target's location in previous frame.

\section{Caltech Roadside Pedestrians (CRP)}
We use CRP for ablation analysis mainly because videos in CRP are long and people are moving very fast, which presents a different tracking scenario comparing to existing dataset including MOT17 and TAO. As CRP is not widely used for multi-person tracking, we adopt the following evaluation protocol: we only evaluate on frames where ground truth is available and we do not penalize detected instances that overlap with background bounding boxes (instance id = 0). As background bounding boxes are not annotated tightly, we enforce a very loose IOU matching, i.e. a detected bounding box is deemed matched to a background one if their IOU overlap is larger then 0.2. 

\paragraph{Training in SiamMOT.} We present the ablation experiments in Tab. \ref{table:ablation_sampling_crp}. Overall, we observe similar trend as that in MOT17, but we don't observe that FP (in MOTA metric) is reduced as significant as in MOT when negative triplets are added ($+\operatorname{N}$) during training. We find this is mainly because 1), detection in CRP is very accurate and 2), CRP is not exhaustively annotated, so large percentage of FP results from tracking un-annotated person in the background rather from real false detection. Note how hard examples (+$\operatorname{H}$) is important to reduce id switches (i.e. false matching).

\begin{table}[t]
    \footnotesize
    \begin{center}
    \begin{tabular}{l|c @{\hskip 0.5em} c @{\hskip 0.5em} c @{\hskip 0.5em} c @{\hskip 0.5em} c @{\hskip 0.5em}}
    \toprule
    \multicolumn{1}{c}{Sampled triplets} & \multicolumn{5}{c}{Caltech Roadside Pedestrians} \\
    \toprule
     & MOTA $\uparrow$ & IDF1 $\uparrow$ & FP $\downarrow$ & FN $\downarrow$ & IDsw $\downarrow$ \\
     \midrule
     P + H & 76.1 & 81.3 & 2679 & 2595 & 1266  \\
     P + N & 74.6 & 79.0 & 2428 & 2768 & 1758 \\
     P + H + N & 76.4 & 81.1 & 2548 & 2575 & 1311 \\
     \bottomrule
    \end{tabular}
    \end{center}
    \caption{\small \it Effects of sampled triplets for training forward tracker in SiamMOT. P / N / H are positive / negative / hard training triplet. P+H triplets are usually used in single-object tracking.
    }
    \label{table:ablation_sampling_crp}
\end{table}

\paragraph{Inference in SiamMOT.} We find that $\tau > 1$ (frame) has negligible effect in CRP. This is mainly because person moves too fast in CRP videos, so the tracker in SiamMOT fails to track them forward beyond 2 frames in CRP. 

\section{MOT17}
We use public detection to generate our results on test set. We follow recent practices \cite{bergmann2019tracking, zhou2020tracking} that re-scores the provided public detection by using the detector in SiamMOT. This is allowed in public detection protocol.  We report detailed video-level metrics in Tab. \ref{table:supplement_results_mot17}.

\begin{table}[t]
    \scriptsize
    \begin{center}
    \begin{tabular}{l @{\hskip 0.25em} c @{\hskip 0.25em} c @{\hskip 0.25em}  c @{\hskip 0.25em}  c @{\hskip 0.5em} c @{\hskip 0.25em} r @{\hskip 0.25em} r @{\hskip 0.25em} r}
    \toprule
    Sequence & Det & MOTA$\uparrow$ & IDF1$\uparrow$ & MT$\uparrow$ & ML$\downarrow$ & FP$\downarrow$ & FN $\downarrow$ & IDsw$\downarrow$ \\
    \midrule
    MOT17-01 & DPM & 53.3 & 47.1 & 33.3\% & 37.5\% & 150 & 2830 & 34 \\
    MOT17-03 & DPM & 76.5	& 71.7 & 57.4\% & 11.5\% & 1359 & 23137 & 131 \\
    MOT17-06 & DPM & 54.9	& 52.7 & 31.9\%	& 30.2\% & 1089	& 4043	& 178 \\
    MOT17-07 & DPM & 59.9	& 52.5 & 23.3\%	& 18.3\% & 651	& 6034	& 86 \\
    MOT17-08 & DPM & 40.1	& 35.1 & 21.1\%	& 31.6\% & 443	& 12094	& 125 \\
    MOT17-12 & DPM & 56.1	& 62.8 & 36.3\%	& 31.9\% & 436  & 3349	& 21 \\
    MOT17-14 & DPM & 43.9	& 49.0 & 15.9\%	& 29.3\% & 947	& 9077	& 340 \\
    \midrule
     MOT17-01 & FRCNN  & 52.5	& 45.6 & 33.3\% & 37.5\% & 198 & 2836 & 27 \\
    MOT17-03 & FRCNN & 76.8	& 74.9 & 56.8\% & 10.1\% & 1428 & 22787 & 123 \\
    MOT17-06 & FRCNN & 58.2	& 54.8 & 37.8\%	& 18.0\% & 1283	& 3412	& 227 \\
    MOT17-07 & FRCNN & 58.2	& 54.0 & 23.3\%	& 15.0\% & 740	& 6264	& 65 \\
    MOT17-08 & FRCNN & 36.4	& 35.5 & 21.1\%	& 39.5\% & 399	& 12933	& 99 \\
    MOT17-12 & FRCNN & 50.1	& 59.2 & 27.5\%	& 41.8\% & 512  & 3796	& 19 \\
    MOT17-14 & FRCNN & 44.2	& 49.7 & 16.5\%	& 28.7\% & 1352	& 8542	& 414 \\
    \midrule
     MOT17-01 & SDP & 55.4	& 47.8 & 33.3\% & 33.3\% & 237 & 2601 & 37 \\
    MOT17-03 & SDP & 82.5	& 74.5 & 68.2\% & 8.10\% & 1846 & 16283 & 183 \\
    MOT17-06 & SDP & 57.6	& 54.7 & 41.0\%	& 23.9\% & 1304	& 3469	& 219 \\
    MOT17-07 & SDP & 62.7	& 52.6 & 33.3\%	& 11.7\% & 984	& 5228	& 89 \\
    MOT17-08 & SDP & 42.1	& 36.7 & 25.0\%	& 28.9\% & 527	& 11559	& 152 \\
    MOT17-12 & SDP & 54.8	& 63.6 & 37.4\%	& 35.2\% & 665  & 3233	& 24 \\
    MOT17-14 & SDP & 48.9	& 63.5 & 18.3\%	& 23.2\% & 1548	& 7448	& 447 \\
    \midrule
    \multicolumn{2}{c}{All} & 65.9 & 63.5 & 34.6\% & 23.9\% & 18098 & 170955 & 3040 \\
    \bottomrule
    \end{tabular}
    \end{center}
    \caption{\small \it Detailed result summary on MOT17 test videos.}
    \label{table:supplement_results_mot17}
\end{table}

\section{HiEve}
We use public detection to generate our results on test videos, the same practice as that in MOT17. Please refer to the following link in official leaderboard for detailed video-level metrics as well as visualized predictions. \url{http://humaninevents.org/tracker.html?tracker=1&id=200}

\section{TAO-person}
\paragraph{Performance per dataset.} We report performance of different subset in TAO-person in Tab. \ref{table:dataset_split_tao}. This dataset-wise performance gives us understanding how SiamMOT performs on different tracking scenarios. Overall, SiamMOT performs very competitive on self-driving street scenes, e.g. BDD and Argoverse as well as on movie dataset Charades. 

\begin{table}[t]
    \scriptsize
    \begin{center}
    \begin{tabular}{l|cc|cc}
    \toprule
    \multicolumn{1}{c}{Subset in TAO} & \multicolumn{2}{c}{SiamMOT(ResNet-101)} & \multicolumn{2}{c}{SiamMOT(DLA-169)} \\
    \toprule
     & TAP@0.5 & TAP@0.75 & TAP@0.5 & TAP@0.75\\
    \midrule
    YFCC100M & 41.3\% & 18.3\% & 40.8\% & 20.0\% \\
    HACS & 33.1\% & 17.3\% & 35.1\% & 18.2\% \\
    BDD & 72.3\% & 41.3\% & 73.8\% & 42.8\% \\
    Argoverse & 66.3\% & 39.5\% & 71.7\% & 42.7\% \\
    AVA & 41.2\% & 25.8\% & 41.8\% & 26.8\% \\
    LaSOT & 28.4\% & 14.9\% & 28.7\% & 16.7\% \\
    Charades & 74.8\% & 68.2\% & 85.7\% & 68.4\% \\
    \midrule
    All & 41.1\% & 23.0\% & 42.1\% & 24.3\% \\
    \bottomrule
    \end{tabular}
    \end{center}
    \caption{\small \it dataset-wise performance on TAO-person.}
    \label{table:dataset_split_tao}
\end{table}

\paragraph{Federated MOTA.} For reference, we also report MOT Challenge metric \cite{milan2016mot16} on Tao-person validation set in Tab. \ref{table:mota_tao}. We find that SiamMOT also significantly outperforms Tracktor++ \cite{dave2020tao} on those metrics.

\begin{table}[htp]
    \scriptsize
    \begin{center}
    \begin{tabular}{l|l@{\hskip 0.5em} c @{\hskip 0.5em} c @{\hskip 0.5em} c @{\hskip 0.5em} c @{\hskip 0.5em} c @{\hskip 0.5em} c @{\hskip 0.5em} c}
    \toprule
     Model & Backbone & MOTA $\uparrow$ & IDF1 $\uparrow$ & MT $\uparrow$ & ML$\downarrow$ & FP $\downarrow$ & FN $\downarrow$ & IDsw $\downarrow$ \\
     \toprule
     Tracktor++ \cite{dave2020tao} & ResNet-101 & 66.6 & 64.8 & 1529 & 411 & 12910 & 2821 & 3487 \\
     \midrule
     SiamMOT & ResNet-101 & 74.6 & 68.0 & 1926 & 204 & 7930 & 4195 & 1816 \\
     SiamMOT & DLA-169 & 75.5 & 68.3 & 1941 & 190 & 7591 & 4176 & 1857 \\
     SiamMOT+ & DLA-169 & 76.7 & 70.9 & 1951 & 190 & 7845 & 3561 & 1834 \\
     \bottomrule
    \end{tabular}
    \end{center}
    \caption{\small \it MOT Challenge metric on TAO-person validation.
    }
    \label{table:mota_tao}
\end{table}


\section{Sensitivity analysis of parameters}
We present the sensitivity analysis of parameters $\alpha$ and $\beta$ that is used in inference, as we observe that the tracking performance is relatively more sensitive to their value changes.  To elaborate, $\alpha$ indicates the detection confidence threshold that we use to start a new trajectory, and $\beta$ is the visibility confidence threshold that is used to determined whether a trajectory needs to be continued. We do a grid search of $\alpha$ ($[0.4: 0.8 : 0.2]$ ) and $\beta$ ($[0.4: 0.8 : 0.2]$), and we present their results on MOT17 in Tab.~\ref{table:alpha_beta}. As expected, large values of $\alpha$ and $\beta$ makes the solver too cautious, which leads to high FN. A good balance is achieved when $\beta=0.4$, and $\alpha=0.6$ is used in the rest of paper to avoid the solver overfitting specifically to MOT17. 

\begin{table}[t]
    \scriptsize
    \begin{center}
    \begin{tabular}{ll|c @{\hskip 0.5em} c @{\hskip 0.5em} c @{\hskip 0.5em} c @{\hskip 0.5em} c @{\hskip 0.5em}}
    \toprule
     $\alpha$& $\beta$ & MOTA $\uparrow$ & IDF1 $\uparrow$ & FP $\downarrow$ & FN $\downarrow$ & IDsw $\downarrow$ \\
     \midrule
     0.4 & 0.4 & 63.8 & 58.5 & 6105 & 33876 & 707  \\
     0.4 & 0.6 & 63.0 & 54.4 & 4973 & 35707 & 922 \\
     0.4 & 0.8 & 59.7 & 51.1 & 2595 & 41686 & 975 \\
     \midrule
     0.6 & 0.4 & 63.3 & 58.4 & 5726 & 34833 & 671  \\
     0.6 & 0.6 & 62.4 & 54.5 & 4330 & 37034 & 869 \\
     0.6 & 0.8 & 59.6 & 51.1 & 2322 & 42167 & 918 \\
     \midrule
     0.8 & 0.4 & 61.8 & 58.3 & 4742 & 37611 & 588  \\
     0.8 & 0.6 & 60.9 & 54.8 & 3169 & 40030 & 729 \\
     0.8 & 0.8 & 58.7 & 51.6 & 1842 & 43730 & 793 \\
     \bottomrule
    \end{tabular}
    \end{center}
    \caption{\small \it Sensitity analysis of $\alpha$ and $\beta$ on MOT17 dataset. The experiment settings are exactly the same as that in ablation analysis.
    }
    \label{table:alpha_beta}
\end{table}
\end{appendices}

{\small
\bibliographystyle{ieee_fullname}
\bibliography{egbib}

\begin{thebibliography}{10}\itemsep=-1pt

\bibitem{andriyenko2011multi}
Anton Andriyenko and Konrad Schindler.
\newblock Multi-target tracking by continuous energy minimization.
\newblock In {\em CVPR 2011}. IEEE, 2011.

\bibitem{ban2016tracking}
Yutong Ban, Sileye Ba, Xavier Alameda-Pineda, and Radu Horaud.
\newblock Tracking multiple persons based on a variational bayesian model.
\newblock In {\em European Conference on Computer Vision}, pages 52--67.
  Springer, 2016.

\bibitem{berclaz2006robust}
Jerome Berclaz, Francois Fleuret, and Pascal Fua.
\newblock Robust people tracking with global trajectory optimization.
\newblock In {\em 2006 IEEE Computer Society Conference on Computer Vision and
  Pattern Recognition (CVPR'06)}. IEEE, 2006.

\bibitem{berclaz2011multiple}
Jerome Berclaz, Francois Fleuret, Engin Turetken, and Pascal Fua.
\newblock Multiple object tracking using k-shortest paths optimization.
\newblock {\em TPAMI}, 33(9):1806--1819, 2011.

\bibitem{bergmann2019tracking}
Philipp Bergmann, Tim Meinhardt, and Laura Leal-Taixe.
\newblock Tracking without bells and whistles.
\newblock In {\em ICCV}, 2019.

\bibitem{bertinetto2016fully}
Luca Bertinetto, Jack Valmadre, Joao~F Henriques, Andrea Vedaldi, and Philip~HS
  Torr.
\newblock Fully-convolutional siamese networks for object tracking.
\newblock In {\em ECCV}, 2016.

\bibitem{bewley2016simple}
Alex Bewley, Zongyuan Ge, Lionel Ott, Fabio Ramos, and Ben Upcroft.
\newblock Simple online and realtime tracking.
\newblock In {\em 2016 IEEE International Conference on Image Processing
  (ICIP)}, pages 3464--3468. IEEE, 2016.

\bibitem{braso2020learning}
Guillem Bras{\'o} and Laura Leal-Taix{\'e}.
\newblock Learning a neural solver for multiple object tracking.
\newblock In {\em Proceedings of the IEEE/CVF Conference on Computer Vision and
  Pattern Recognition}, pages 6247--6257, 2020.

\bibitem{Argoverse}
Ming-Fang Chang, John~W Lambert, Patsorn Sangkloy, Jagjeet Singh, Slawomir Bak,
  Andrew Hartnett, De Wang, Peter Carr, Simon Lucey, Deva Ramanan, and James
  Hays.
\newblock Argoverse: 3d tracking and forecasting with rich maps.
\newblock In {\em Conference on Computer Vision and Pattern Recognition
  (CVPR)}, 2019.

\bibitem{choi2015near}
Wongun Choi.
\newblock Near-online multi-target tracking with aggregated local flow
  descriptor.
\newblock In {\em Proceedings of the IEEE international conference on computer
  vision}, 2015.

\bibitem{choi2010multiple}
Wongun Choi and Silvio Savarese.
\newblock Multiple target tracking in world coordinate with single, minimally
  calibrated camera.
\newblock In {\em European Conference on Computer Vision}. Springer, 2010.

\bibitem{dai2016r}
Jifeng Dai, Yi Li, Kaiming He, and Jian Sun.
\newblock R-fcn: Object detection via region-based fully convolutional
  networks.
\newblock In {\em NeurIPS}, 2016.

\bibitem{danelljan2019atom}
Martin Danelljan, Goutam Bhat, Fahad~Shahbaz Khan, and Michael Felsberg.
\newblock Atom: Accurate tracking by overlap maximization.
\newblock In {\em Proceedings of the IEEE Conference on Computer Vision and
  Pattern Recognition}, pages 4660--4669, 2019.

\bibitem{dave2020tao}
Achal Dave, Tarasha Khurana, Pavel Tokmakov, Cordelia Schmid, and Deva Ramanan.
\newblock Tao: A large-scale benchmark for tracking any object.
\newblock In {\em European Conference on Computer Vision}, 2020.

\bibitem{dosovitskiy2015flownet}
Alexey Dosovitskiy, Philipp Fischer, Eddy Ilg, Philip Hausser, Caner Hazirbas,
  Vladimir Golkov, Patrick Van Der~Smagt, Daniel Cremers, and Thomas Brox.
\newblock Flownet: Learning optical flow with convolutional networks.
\newblock In {\em Proceedings of the IEEE international conference on computer
  vision}, pages 2758--2766, 2015.

\bibitem{duan2019centernet}
Kaiwen Duan, Song Bai, Lingxi Xie, Honggang Qi, Qingming Huang, and Qi Tian.
\newblock Centernet: Keypoint triplets for object detection.
\newblock In {\em Proceedings of the IEEE International Conference on Computer
  Vision}, pages 6569--6578, 2019.

\bibitem{evangelidis2008parametric}
Georgios~D Evangelidis and Emmanouil~Z Psarakis.
\newblock Parametric image alignment using enhanced correlation coefficient
  maximization.
\newblock {\em IEEE Transactions on Pattern Analysis and Machine Intelligence},
  30(10):1858--1865, 2008.

\bibitem{fan2019siamese}
Heng Fan and Haibin Ling.
\newblock Siamese cascaded region proposal networks for real-time visual
  tracking.
\newblock In {\em CVPR}, 2019.

\bibitem{fang2018recurrent}
Kuan Fang, Yu Xiang, Xiaocheng Li, and Silvio Savarese.
\newblock Recurrent autoregressive networks for online multi-object tracking.
\newblock In {\em 2018 IEEE Winter Conference on Applications of Computer
  Vision (WACV)}, pages 466--475. IEEE, 2018.

\bibitem{girshick2015fast}
Ross Girshick.
\newblock Fast r-cnn.
\newblock In {\em ICCV}, 2015.

\bibitem{gu2018ava}
Chunhui Gu, Chen Sun, David~A Ross, Carl Vondrick, Caroline Pantofaru, Yeqing
  Li, Sudheendra Vijayanarasimhan, George Toderici, Susanna Ricco, Rahul
  Sukthankar, et~al.
\newblock Ava: A video dataset of spatio-temporally localized atomic visual
  actions.
\newblock In {\em Proceedings of the IEEE Conference on Computer Vision and
  Pattern Recognition}, pages 6047--6056, 2018.

\bibitem{guo2020siamcar}
Dongyan Guo, Jun Wang, Ying Cui, Zhenhua Wang, and Shengyong Chen.
\newblock Siamcar: Siamese fully convolutional classification and regression
  for visual tracking.
\newblock In {\em Proceedings of the IEEE/CVF Conference on Computer Vision and
  Pattern Recognition}, pages 6269--6277, 2020.

\bibitem{guo2017learning}
Qing Guo, Wei Feng, Ce Zhou, Rui Huang, Liang Wan, and Song Wang.
\newblock Learning dynamic siamese network for visual object tracking.
\newblock In {\em ICCV}, 2017.

\bibitem{gupta2019lvis}
Agrim Gupta, Piotr Dollar, and Ross Girshick.
\newblock Lvis: A dataset for large vocabulary instance segmentation.
\newblock In {\em Proceedings of the IEEE Conference on Computer Vision and
  Pattern Recognition}, pages 5356--5364, 2019.

\bibitem{hall2015fine}
David Hall and Pietro Perona.
\newblock Fine-grained classification of pedestrians in video: Benchmark and
  state of the art.
\newblock In {\em Proceedings of the IEEE Conference on Computer Vision and
  Pattern Recognition}, pages 5482--5491, 2015.

\bibitem{he2018twofold}
Anfeng He, Chong Luo, Xinmei Tian, and Wenjun Zeng.
\newblock A twofold siamese network for real-time object tracking.
\newblock In {\em CVPR}, 2018.

\bibitem{he2017mask}
Kaiming He, Georgia Gkioxari, Piotr Doll{\'a}r, and Ross Girshick.
\newblock Mask r-cnn.
\newblock In {\em ICCV}, 2017.

\bibitem{held2016learning}
David Held, Sebastian Thrun, and Silvio Savarese.
\newblock Learning to track at 100 fps with deep regression networks.
\newblock In {\em ECCV}, 2016.

\bibitem{henschel2017improvements}
Roberto Henschel, Laura Leal-Taix{\'e}, Daniel Cremers, and Bodo Rosenhahn.
\newblock Improvements to frank-wolfe optimization for multi-detector
  multi-object tracking.
\newblock {\em arXiv preprint arXiv:1705.08314}, 2017.

\bibitem{hermans2017defense}
Alexander Hermans, Lucas Beyer, and Bastian Leibe.
\newblock In defense of the triplet loss for person re-identification.
\newblock In {\em CVPRW}, 2017.

\bibitem{kalman1960new}
Rudolph~Emil Kalman.
\newblock A new approach to linear filtering and prediction problems.
\newblock 1960.

\bibitem{keuper2018motion}
Margret Keuper, Siyu Tang, Bjoern Andres, Thomas Brox, and Bernt Schiele.
\newblock Motion segmentation \& multiple object tracking by correlation
  co-clustering.
\newblock {\em TPAMI}, 42(1):140--153, 2018.

\bibitem{kim2015multiple}
Chanho Kim, Fuxin Li, Arridhana Ciptadi, and James~M Rehg.
\newblock Multiple hypothesis tracking revisited.
\newblock In {\em ICCV}, 2015.

\bibitem{leal2016learning}
Laura Leal-Taix{\'e}, Cristian Canton-Ferrer, and Konrad Schindler.
\newblock Learning by tracking: Siamese cnn for robust target association.
\newblock In {\em Proceedings of the IEEE Conference on Computer Vision and
  Pattern Recognition Workshops}, 2016.

\bibitem{li2019siamrpn++}
Bo Li, Wei Wu, Qiang Wang, Fangyi Zhang, Junliang Xing, and Junjie Yan.
\newblock Siamrpn++: Evolution of siamese visual tracking with very deep
  networks.
\newblock In {\em CVPR}, 2019.

\bibitem{li2018high}
Bo Li, Junjie Yan, Wei Wu, Zheng Zhu, and Xiaolin Hu.
\newblock High performance visual tracking with siamese region proposal
  network.
\newblock In {\em CVPR}, 2018.

\bibitem{lin2017feature}
Tsung-Yi Lin, Piotr Doll{\'a}r, Ross Girshick, Kaiming He, Bharath Hariharan,
  and Serge Belongie.
\newblock Feature pyramid networks for object detection.
\newblock In {\em CVPR}, 2017.

\bibitem{lin2017focal}
Tsung-Yi Lin, Priya Goyal, Ross Girshick, Kaiming He, and Piotr Doll{\'a}r.
\newblock Focal loss for dense object detection.
\newblock In {\em Proceedings of the IEEE international conference on computer
  vision}, pages 2980--2988, 2017.

\bibitem{lin2014microsoft}
Tsung-Yi Lin, Michael Maire, Serge Belongie, James Hays, Pietro Perona, Deva
  Ramanan, Piotr Doll{\'a}r, and C~Lawrence Zitnick.
\newblock Microsoft coco: Common objects in context.
\newblock In {\em ECCV}, 2014.

\bibitem{lin2020hieve}
Weiyao Lin, Huabin Liu, Shizhan Liu, Yuxi Li, Guo-Jun Qi, Rui Qian, Tao Wang,
  Nicu Sebe, Ning Xu, Hongkai Xiong, et~al.
\newblock {ACM MM Grand Challenge on Large-scale Human-centric Video Analysis
  in Complex Events}.
\newblock \url{http://humaninevents.org/ACM_welcome.html}, 2020.

\bibitem{lin2020human}
Weiyao Lin, Huabin Liu, Shizhan Liu, Yuxi Li, Guo-Jun Qi, Rui Qian, Tao Wang,
  Nicu Sebe, Ning Xu, Hongkai Xiong, et~al.
\newblock Human in events: A large-scale benchmark for human-centric video
  analysis in complex events.
\newblock {\em arXiv preprint arXiv:2005.04490}, 2020.

\bibitem{milan2016mot16}
Anton Milan, Laura Leal-Taix{\'e}, Ian Reid, Stefan Roth, and Konrad Schindler.
\newblock Mot16: A benchmark for multi-object tracking.
\newblock {\em arXiv preprint arXiv:1603.00831}, 2016.

\bibitem{peng2020dense}
Jinlong Peng, Yueyang Gu, Yabiao Wang, Chengjie Wang, Jilin Li, and Feiyue
  Huang.
\newblock Dense scene multiple object tracking with box-plane matching.
\newblock In {\em Proceedings of the 28th ACM International Conference on
  Multimedia}, pages 4615--4619, 2020.

\bibitem{qiu2020borderdet}
Han Qiu, Yuchen Ma, Zeming Li, Songtao Liu, and Jian Sun.
\newblock Borderdet: Border feature for dense object detection.
\newblock In {\em European Conference on Computer Vision}. Springer, 2020.

\bibitem{ren2015faster}
Shaoqing Ren, Kaiming He, Ross Girshick, and Jian Sun.
\newblock Faster r-cnn: Towards real-time object detection with region proposal
  networks.
\newblock In {\em NeurIPS}, 2015.

\bibitem{ristani2018features}
Ergys Ristani and Carlo Tomasi.
\newblock Features for multi-target multi-camera tracking and
  re-identification.
\newblock In {\em CVPR}, 2018.

\bibitem{sadeghian2017tracking}
Amir Sadeghian, Alexandre Alahi, and Silvio Savarese.
\newblock Tracking the untrackable: Learning to track multiple cues with
  long-term dependencies.
\newblock In {\em Proceedings of the IEEE International Conference on Computer
  Vision}, 2017.

\bibitem{shao2018crowdhuman}
Shuai Shao, Zijian Zhao, Boxun Li, Tete Xiao, Gang Yu, Xiangyu Zhang, and Jian
  Sun.
\newblock Crowdhuman: A benchmark for detecting human in a crowd.
\newblock {\em arXiv preprint arXiv:1805.00123}, 2018.

\bibitem{sheng2018heterogeneous}
Hao Sheng, Yang Zhang, Jiahui Chen, Zhang Xiong, and Jun Zhang.
\newblock Heterogeneous association graph fusion for target association in
  multiple object tracking.
\newblock {\em TCSVT}, 29(11):3269--3280, 2018.

\bibitem{sigurdsson2016hollywood}
Gunnar~A Sigurdsson, G{\"u}l Varol, Xiaolong Wang, Ali Farhadi, Ivan Laptev,
  and Abhinav Gupta.
\newblock Hollywood in homes: Crowdsourcing data collection for activity
  understanding.
\newblock In {\em European Conference on Computer Vision}, pages 510--526.
  Springer, 2016.

\bibitem{sun2018pwc}
Deqing Sun, Xiaodong Yang, Ming-Yu Liu, and Jan Kautz.
\newblock Pwc-net: Cnns for optical flow using pyramid, warping, and cost
  volume.
\newblock In {\em CVPR}, 2018.

\bibitem{sun2019deep}
ShiJie Sun, Naveed Akhtar, HuanSheng Song, Ajmal Mian, and Mubarak Shah.
\newblock Deep affinity network for multiple object tracking.
\newblock {\em IEEE transactions on pattern analysis and machine intelligence},
  43(1):104--119, 2019.

\bibitem{tang2017multiple}
Siyu Tang, Mykhaylo Andriluka, Bjoern Andres, and Bernt Schiele.
\newblock Multiple people tracking by lifted multicut and person
  re-identification.
\newblock In {\em CVPR}, 2017.

\bibitem{tao2016siamese}
Ran Tao, Efstratios Gavves, and Arnold~WM Smeulders.
\newblock Siamese instance search for tracking.
\newblock In {\em CVPR}, 2016.

\bibitem{tian2019directpose}
Zhi Tian, Hao Chen, and Chunhua Shen.
\newblock Directpose: Direct end-to-end multi-person pose estimation.
\newblock In {\em Proceedings of the IEEE conference on computer vision and
  pattern recognition}, 2020.

\bibitem{tian2019fcos}
Zhi Tian, Chunhua Shen, Hao Chen, and Tong He.
\newblock Fcos: Fully convolutional one-stage object detection.
\newblock In {\em Proceedings of the IEEE international conference on computer
  vision}, pages 9627--9636, 2019.

\bibitem{valmadre2017end}
Jack Valmadre, Luca Bertinetto, Joao Henriques, Andrea Vedaldi, and Philip~HS
  Torr.
\newblock End-to-end representation learning for correlation filter based
  tracking.
\newblock In {\em CVPR}, 2017.

\bibitem{varior2016siamese}
Rahul~Rama Varior, Bing Shuai, Jiwen Lu, Dong Xu, and Gang Wang.
\newblock A siamese long short-term memory architecture for human
  re-identification.
\newblock In {\em European conference on computer vision}, pages 135--153.
  Springer, 2016.

\bibitem{wang2019exploit}
Gaoang Wang, Yizhou Wang, Haotian Zhang, Renshu Gu, and Jenq-Neng Hwang.
\newblock Exploit the connectivity: Multi-object tracking with trackletnet.
\newblock In {\em Proceedings of the 27th ACM International Conference on
  Multimedia}, 2019.

\bibitem{wojke2017simple}
Nicolai Wojke, Alex Bewley, and Dietrich Paulus.
\newblock Simple online and realtime tracking with a deep association metric.
\newblock In {\em 2017 IEEE international conference on image processing
  (ICIP)}, pages 3645--3649. IEEE, 2017.

\bibitem{wu2020transductive}
Ancong Wu, Chengzhi Lin, Bogao Chen, Weihao Huang, Zeyu Huang, and Wei-Shi
  Zheng.
\newblock Transductive multi-object tracking in complex events by interactive
  self-training.
\newblock In {\em Proceedings of the 28th ACM International Conference on
  Multimedia}, pages 4620--4624, 2020.

\bibitem{wu2018group}
Yuxin Wu and Kaiming He.
\newblock Group normalization.
\newblock In {\em Proceedings of the European conference on computer vision
  (ECCV)}, pages 3--19, 2018.

\bibitem{xiang2015learning}
Yu Xiang, Alexandre Alahi, and Silvio Savarese.
\newblock Learning to track: Online multi-object tracking by decision making.
\newblock In {\em Proceedings of the IEEE international conference on computer
  vision}, pages 4705--4713, 2015.

\bibitem{xu2019spatial}
Jiarui Xu, Yue Cao, Zheng Zhang, and Han Hu.
\newblock Spatial-temporal relation networks for multi-object tracking.
\newblock In {\em ICCV}, 2019.

\bibitem{xu2019train}
Yihong Xu, Aljosa Osep, Yutong Ban, Radu Horaud, Laura Leal-Taixe, and Xavier
  Alameda-Pineda.
\newblock How to train your deep multi-object tracker.
\newblock {\em arXiv preprint arXiv:1906.06618}, 2019.

\bibitem{yang2019video}
Linjie Yang, Yuchen Fan, and Ning Xu.
\newblock Video instance segmentation.
\newblock In {\em ICCV}, 2019.

\bibitem{yang2019reppoints}
Ze Yang, Shaohui Liu, Han Hu, Liwei Wang, and Stephen Lin.
\newblock Reppoints: Point set representation for object detection.
\newblock In {\em Proceedings of the IEEE International Conference on Computer
  Vision}, pages 9657--9666, 2019.

\bibitem{yu2020bdd100k}
Fisher Yu, Haofeng Chen, Xin Wang, Wenqi Xian, Yingying Chen, Fangchen Liu,
  Vashisht Madhavan, and Trevor Darrell.
\newblock Bdd100k: A diverse driving dataset for heterogeneous multitask
  learning.
\newblock In {\em Proceedings of the IEEE/CVF Conference on Computer Vision and
  Pattern Recognition}, pages 2636--2645, 2020.

\bibitem{yu2018deep}
Fisher Yu, Dequan Wang, Evan Shelhamer, and Trevor Darrell.
\newblock Deep layer aggregation.
\newblock In {\em Proceedings of the IEEE conference on computer vision and
  pattern recognition}, pages 2403--2412, 2018.

\bibitem{yu2016unitbox}
Jiahui Yu, Yuning Jiang, Zhangyang Wang, Zhimin Cao, and Thomas Huang.
\newblock Unitbox: An advanced object detection network.
\newblock In {\em Proceedings of the 24th ACM international conference on
  Multimedia}, pages 516--520, 2016.

\bibitem{zamir2012gmcp}
Amir~Roshan Zamir, Afshin Dehghan, and Mubarak Shah.
\newblock Gmcp-tracker: Global multi-object tracking using generalized minimum
  clique graphs.
\newblock In {\em ECCV}, 2012.

\bibitem{zhang2008global}
Li Zhang, Yuan Li, and Ramakant Nevatia.
\newblock Global data association for multi-object tracking using network
  flows.
\newblock In {\em CVPR}, 2008.

\bibitem{zhang2019structured}
Yubo Zhang, Pavel Tokmakov, Martial Hebert, and Cordelia Schmid.
\newblock A structured model for action detection.
\newblock In {\em CVPR}, 2019.

\bibitem{zhang2019deeper}
Zhipeng Zhang and Houwen Peng.
\newblock Deeper and wider siamese networks for real-time visual tracking.
\newblock In {\em CVPR}, 2019.

\bibitem{zhou2019deep}
Fengwei Zhou, Bin Wu, and Zhenguo Li.
\newblock Deep meta-learning: Learning to learn in the concept space.
\newblock In {\em ICCV}, 2019.

\bibitem{zhou2020tracking}
Xingyi Zhou, Vladlen Koltun, and Philipp Kr{\"a}henb{\"u}hl.
\newblock Tracking objects as points.
\newblock {\em ECCV}, 2020.

\bibitem{zhu2019soft}
Chenchen Zhu, Fangyi Chen, Zhiqiang Shen, and Marios Savvides.
\newblock Soft anchor-point object detection.
\newblock In {\em European Conference on Computer Vision}. Springer, 2020.

\bibitem{zhu2018distractor}
Zheng Zhu, Qiang Wang, Bo Li, Wei Wu, Junjie Yan, and Weiming Hu.
\newblock Distractor-aware siamese networks for visual object tracking.
\newblock In {\em ECCV}, 2018.

\end{thebibliography}
}

\end{document}